\documentclass{article}

    \PassOptionsToPackage{numbers, compress}{natbib}

 \usepackage[eandd, final]{neurips_2026}

\usepackage[utf8]{inputenc} 
\usepackage[T1]{fontenc}    
\usepackage{hyperref}       
\usepackage{url}            
\usepackage{booktabs}       
\usepackage{amsfonts}       
\usepackage{nicefrac}       
\usepackage{microtype}      
\usepackage{xcolor}         
\usepackage{graphicx}
\usepackage{amsmath}
\usepackage{multirow}
\usepackage{makecell}
\usepackage{caption}
\usepackage{colortbl}
\usepackage{tcolorbox}
\usepackage[tikz]{bclogo}
\usepackage{subfigure}
\usepackage{subcaption}
\usepackage{arydshln}
\usepackage{float}
\usepackage{pifont}
\usepackage{alltt}
\usepackage{dblfloatfix}
\usepackage{enumitem}
\usepackage{wrapfig}
\title{Towards Real-world Human Behavior Simulation: Benchmarking Large Language Models on Long-horizon, Cross-scenario, Heterogeneous Behavior Traces}

\author{
Jiawei Chen$^{1,2}$\thanks{Equal contribution. This work was done when Jiawei and Ruoxi interned at Kuaishou.}\hspace{0.5em},
Ruoxi Xu$^{1,2}$\footnotemark[1]\hspace{0.5em},
Boxi Cao$^1$\thanks{Corresponding authors.}\hspace{0.5em},
Ruotong Pan$^3$,
Yunfei Zhang$^3$,
Yifei Hu$^3$,\\
\textbf{Yong Du}$^3$,
\textbf{Tingting Gao}$^3$,
\textbf{Han Li}$^3$,
\textbf{Yaojie Lu}$^1$,
\textbf{Yingfei Sun}$^2$,\\
\textbf{Xianpei Han}$^1$,
\textbf{Le Sun}$^1$,
\textbf{Xiangyu Wu}$^3$\footnotemark[2]\hspace{0.5em},
\textbf{Hongyu Lin}$^1$\\
$^1$ Chinese Information Processing Laboratory, Institute of Software, \\
  Chinese Academy of Sciences \\
$^2$ University of Chinese Academy of Sciences \\
$^3$ Kuaishou Technology \\[3pt]
\url{https://OmniBehavior.github.io}
} 
\hypersetup{
    colorlinks=true,
    linkcolor=black,
    citecolor=black,
    urlcolor=blue
}

\tcbuselibrary{skins}
\definecolor{msftBlack}{RGB}{0,0,0}
\newtcolorbox{findingBox}{
    enhanced,
    colback=msftBlack!05,
    colframe=msftBlack!10,
    arc=1mm,
    boxrule=0.5pt,
    left=2mm, right=2mm, top=2mm, bottom=2mm,
    drop fuzzy shadow,
    fontupper=\em,
    notitle
}

\newcommand{\cmark}{\ding{51}} 
\newcommand{\xmark}{\ding{55}} 

\begin{document}

\maketitle
\setcounter{footnote}{0}

\vspace{-0.5em}
\begin{abstract}
The emergence of Large Language Models (LLMs) has illuminated the potential for a general-purpose user simulator. However, existing benchmarks remain constrained to isolated scenarios, narrow action spaces, or synthetic data, failing to capture the holistic nature of authentic human behavior. 
To bridge this gap, we introduce \textbf{\textit{OmniBehavior}}, the first predictive user simulation benchmark constructed entirely from real-world data, integrating long-horizon, cross-scenario, and heterogeneous behavioral patterns into a unified framework. Based on this benchmark, we first provide empirical evidence that previous datasets with isolated scenarios suffer from tunnel vision, whereas real-world decision-making relies on long-term, cross-scenario causal chains. Extensive evaluations of state-of-the-art LLMs reveal that current models struggle to accurately predict these complex behaviors, with performance plateauing even as context windows expand. Crucially, a systematic comparison between simulated and authentic behaviors uncovers a fundamental structural bias: LLMs tend to converge towards a positive average person, exhibiting hyper-activity, persona homogenization, and a utopian bias. This results in the loss of individual differences and long-tail behaviors, highlighting critical directions for future high-fidelity simulation research\footnote{We have released the code on \href{https://github.com/icip-cas/OmniBehavior}{GitHub} and the dataset on \href{https://huggingface.co/datasets/jiawei-ucas/OmniBehavior}{Hugging Face}, with Chinese and English versions available.}.
\end{abstract}

\section{Introduction}

Human behavior is not a sequence of isolated atoms, but a continuous narrative traversing long horizons and diverse scenarios~\citep{lewin1951field}. The holistic modeling of such behavior is central to a wide range of disciplines, such as cognitive science~\citep{naveed2019cognitive, mcclelland2009place},  behavioral economics~\citep{arthur1991designing, clarkson1960simulation, reiss2011plea}, recommender systems~\citep{zhang2020evaluating, bougie2025simuser, bernardi2021simulations}, and interactive artificial intelligence~\citep{lavery1986artificial, petrovic2018artificial}. 
The recent breakthrough advancements of Large Language Models (LLMs) have raised an appealing prospect that a single LLM might serve as a general-purpose user simulator: predicting personalized behavior, and standing in for authentic users in the design and evaluation of interactive systems. To succeed, such a model must faithfully portray the cross-scenario, long-horizon, and heterogeneous structure of behavior, rather than merely fitting short, homogeneous interaction segments.

While LLM-based simulators are rapidly advancing~\citep{wangsurvey, gao2024large, taillandier2025integrating}, the empirical basis for assessing their capabilities remains systematically insufficient.
Specifically, existing benchmarks are often confined to isolated scenarios (e.g., exclusively focusing on video browsing~\citep{gao2022kuairec}, live streaming~\citep{qu2025kuailive}, E-commerce~\citep{wang2026opera}, customer service dialogue~\citep{zhao2021jddc}) with narrow action spaces (e.g., click or watch).
\emph{Such narrow focus overlooks the holistic nature of real human behavior, where authentic user preferences and actions are intrinsically interconnected across various scenarios and activity types.}
A user's comprehensive digital footprint constitutes a cross-scenario continuum, composed of heterogeneous behaviors ranging from passive browsing and clicking to active searching, commenting, consulting, purchasing, etc. 
For instance, a purchase decision of a user may stem from interest sparked by a video viewed days earlier, while subsequent comments in a live stream reflect the actual post-purchase experience. 
This reliance on fragmented evaluation undermines the reliable evaluation of LLMs’ behavioral modeling capacity and risks systematic misinterpretation, thereby constraining both optimization and real-world deployment.

\begin{figure*}[t]
\centering
\includegraphics[width=\textwidth]{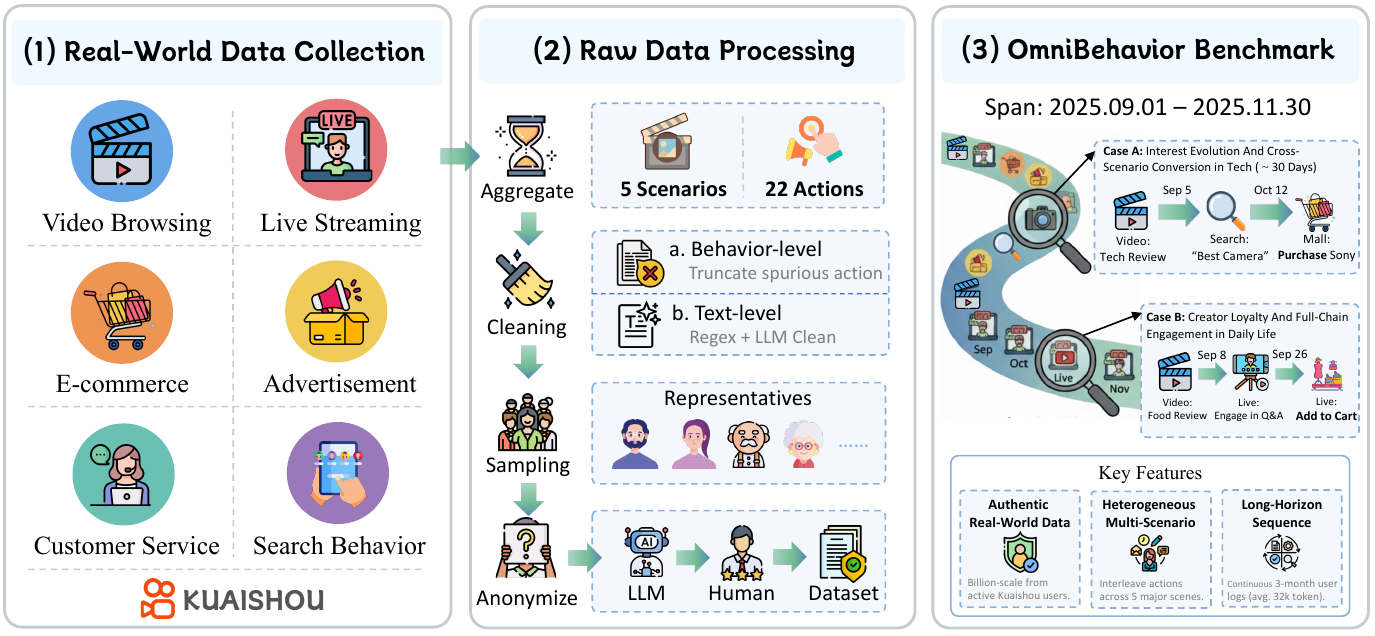}
\caption{Overview of \textbf{OmniBehavior}, a real-world comprehensive benchmark for evaluating LLM-based user simulators. The benchmark is constructed in three stages: \textbf{(1) Data Collection}: aggregation of real-world logs from the Kuaishou platform across several major scenarios, with \emph{Customer Service} treated as part of the \emph{E-commerce} scenario, yielding 5 scenarios in total after aggregation. \textbf{(2) Data Processing}: multi-modal fusion, two-level cleaning, representative sampling, and anonymization. \textbf{(3) Benchmark Construction}: the resulting dataset captures long-horizon, cross-scenario behavior traces, providing a high-fidelity testbed for evaluating LLM-based user simulators in real-world industrial settings.}
\label{fig:framework}
\vspace{-1.8em}
\end{figure*}

To bridge this gap, \textbf{we propose OmniBehavior, to our knowledge, the first predictive user simulation benchmark built entirely on real-world data that simultaneously captures long-horizon, cross-scenario and heterogeneous behavioral patterns.}
OmniBehavior is collected from Kuaishou Technology\footnote{www.kuaishou.com/en}, one of the world's largest video platforms, with over 400 million daily active users. 
As illustrated in Figure~\ref{fig:framework}, users on Kuaishou navigate across multiple scenarios such as video browsing, live streaming, advertising, and e-commerce, and interact through heterogeneous actions including searches, likes, shares, purchase and conversations. 
The combination of scenario diversity and richly structured interaction traces makes Kuaishou a uniquely suitable source for modeling real-world user behavior.
Based on real user data from Kuaishou, we construct OmniBehavior through careful user sampling, long-horizon behavior aggregation, rigorous quality control, and strict anonymization procedures.
Finally, we aggregate complete interaction traces with precise timestamps from 200 representative users over a three-month period, purposely selected to ensure high trajectory density and diversity, yielding a total of 2.12 million interactions. 
The trace lengths range from 50 to over 100,000, covering a total of 22 distinct actions across 5 different scenarios. 
To evaluate the ability of LLMs to perform predictive user simulation, we formulate a user-conditioned prediction task: given a user profile, the user’s historical behavior sequence, and the context of a specific scenario (e.g., video content, product information, or consultation background), the LLM is required to predict all corresponding user behaviors in that scenario (e.g., purchases and conversational responses).

Drawing on OmniBehavior, we first conduct an in-depth investigation into real users’ behavioral patterns, and reveal that \textbf{cross-scenario dependencies and long-horizon heterogeneous data are essential to transcending fragmented snapshots, enabling a panoramic understanding of user preferences and the modeling of truly authentic behavioral trajectories.}
Specifically, we find that (1) \textit{Single-scenario data suffers from tunnel vision.} Statistically, integrating new scenarios consistently expands interest coverage by approximately 20-30\%, merging disjointed facets into a holistic view that eliminates profiling bias. \textit{(2) Human decision-making inherently follows a long-horizon, cross-scenario causal structure.} Our backtracking analysis reveals that over 80\% of conversion paths span multiple scenarios and extend over days. This confirms that long-sequence, heterogeneous data is essential for modeling authentic trajectories and preserving causal integrity. \textit{(3) Synthetic datasets exhibit significant statistical deviations when compared to authentic human data}, underscoring the irreplaceable value of benchmarks grounded in real-world data.
Given the profound complexity of these authentic patterns, it remains an open question whether state-of-the-art models can transcend fragmented data to capture holistic human decision-making. This necessitates a systematic evaluation of their true behavioral modeling capacities.

Building upon this benchmark, we perform a comprehensive evaluation of current state-of-the-art LLMs, including both closed-source models (e.g., Claude-4.5-Opus~\citep{anthropic2025claudeopus45}, GPT-5.2~\citep{gpt5}) and open-source models (e.g., GLM-4.7~\citep{zai2025glm47}, Qwen3-235B~\citep{yang2025qwen3}), and find that current LLMs exhibit substantial limitations in faithfully modeling long-horizon, cross-scenario user behavior.
For instance, even the best-performing LLM, Claude-4.5-Opus, achieves only an overall score of 44.55. Moreover, the F1 scores on binary behavior prediction tasks (e.g., like and share) of most models do not exceed 40\%.
Furthermore, although existing LLMs support long context windows, we observe that when the context is extended beyond 32K or even to 128K tokens, simply providing longer input sequences does not necessarily yield better performance. This finding highlights the importance of improving long-context reasoning capabilities and designing more effective memory management strategies.

Crucially, we further conduct a systematic comparison between simulated and authentic behaviors, uncovering a fundamental structural bias in current LLM-based simulators: \textbf{Current LLMs tend to converge towards a positive average person representation, blurring individual-specific differences and discarding long-tail behaviors as well as negative interaction patterns, which severely limits their ability to accurately model real human behavior.}
Specifically, we first observe hyper-activity: LLMs consistently overestimate action probabilities compared to human baselines. This hampers their ability to model negative feedback signals, which are crucial for churn prediction.
This quantitative error further leads to persona homogenization: quantitative analysis of behavioral vectors reveals that while real data exhibits distinct separation between individuals, simulated populations show severe overlap. This convergence towards an "average person" implies that models might discard specific, long-tail behavioral patterns in favor of generic responses, failing to reproduce the heterogeneity of real ecosystems.
Finally, simulations are further distorted by a utopian bias: influenced by alignment mechanisms, LLMs are unrealistically positive and polite, and struggle to simulate the adversarial interactions of dissatisfied users.
Collectively, these findings reveal that current LLMs do not merely fall short in accuracy, but exhibit a structural ``positivity-and-average'' filter that prevents the modeling of authentic human diversity.
Our work provides a critical cautionary framework for the use of AI in social science and behavioral modeling, suggesting that without addressing these intrinsic distortions, LLMs may remain mirrors of our ideals rather than maps of our reality.

The major contributions of this paper are summarized as follows:
\begin{itemize}[leftmargin=2em, itemsep=0pt, topsep=0pt]
\item We introduce OmniBehavior, to our knowledge, the first predictive user simulation benchmark constructed entirely from authentic user interaction logs, integrating long-horizon, cross-scenario and heterogeneous behavior traces into a unified framework.
\item We provide the systematic analysis of real-world user behavior at scale, demonstrating that cross-scenario dependencies, long-horizon structures and heterogeneous signals are fundamental to accurate preference modeling.
\item We conduct a comprehensive evaluation of SOTA LLMs, revealing substantial capability gaps in modeling realistic user behavior, even with extended context lengths, and establishing strong baselines for future research.
\item We reveal a structural bias in LLM-based simulators, termed \emph{positivity-and-average bias}, where models overestimate engagement, homogenize user behaviors, and suppress negative and long-tail interactions, fundamentally limiting their applicability in real-world settings.
\end{itemize}

\section{OmniBehavior Benchmark}

In this section, we introduce the framework and construction of OmniBehavior. 
Unlike fragmented datasets, OmniBehavior is built upon end-to-end industrial logs from Kuaishou, introducing three interconnected challenges for LLMs.
First, by covering all major scenarios within Kuaishou, it unifies diverse activities, requiring models to trace causal links across scenarios. Second, by aggregating complete three-month interaction histories, it necessitates modeling ultra-long behavioral sequences. Third, being built entirely on real user traces, it ensures simulations reflect authentic human logic.
By faithfully preserving these structural properties, OmniBehavior establishes a rigorous standard for determining whether a user simulator can serve as a credible and robust proxy for real human behavior in industrial applications.

\subsection{Taxonomy}
\label{sec:framework}

We derive the benchmark taxonomy encompassing 5 major scenarios and 22 distinct user actions on Kuaishou platform (one of the world's largest video platforms). This framework ensures that the benchmark can model the holistic nature of user decision-making:

\begin{itemize}[leftmargin=2em, itemsep=-1pt, topsep=0pt]
    \item \textbf{Video Browsing:} A fundamental content consumption scenario. User behaviors include passive consumption (e.g., \textit{Watch}, \textit{Completion}) and active feedback (e.g., \textit{Like}, \textit{Collect}, \textit{Share}, \textit{Comment}, \textit{Download}, \textit{Follow}, \textit{Dislike}, \textit{Unfollow}).
    \item \textbf{Live Streaming:} A live broadcast scenario focused on real-time interaction. User behaviors span social engagement (e.g., \textit{Like}, \textit{Comment}, \textit{Gift}), content governance (e.g., \textit{Report}), and commerce initiation (e.g., \textit{Add-to-Cart}, \textit{Purchase}).
    \item \textbf{Advertisement:} A commercial scenario interleaved with feeds. Beyond standard feedbacks in video scenario, it involves conversion-specific actions including \textit{Click}, \textit{Form Submission}, \textit{Activation}, and \textit{Purchase}.
    \item \textbf{E-commerce:} An E-commerce scenario focusing on transactional intent. It consists of decision-oriented actions (e.g. \textit{Add-to-Cart}, \textit{Purchase}) and customer service interactions, logging multi-turn dialogues for product inquiries and after-sales service.
    \item \textbf{Search Behavior:} An information retrieval scenario driven by explicit user intent, where users proactively express their needs through targeted search queries within the App.
\end{itemize}

\subsection{Construction Pipeline}
\label{sec:construction_pipeline}

This section presents the end-to-end construction pipeline of our benchmark, including data collection, cleaning, sampling, and anonymization. The pipeline is carefully designed to ensure data quality, representativeness, and compliance with privacy and ethical standards. By integrating multiple processing stages, we aim to construct high-fidelity user behavior sequences that accurately reflect real-world decision-making processes.

\noindent\textbf{Data Collection.}
We collect interaction logs from the Kuaishou platform over a three-month period from September 1 to November 30, 2025. To preserve the holistic structure of user decision-making, we aggregate raw logs from the scenarios defined in Section~\ref{sec:framework} and interleave them based on timestamps to construct unified, chronological behavioral sequences. For each item, we retrieve rich metadata, including content features (e.g., captions, price, OCR, ASR), inherent item attributes (e.g., video duration, commodity description), and interaction statistics (e.g., view counts, paid counts).

\noindent\textbf{Data Cleaning.}
We conduct a two-level cleaning pipeline to systematically eliminate noise:
(1) Behavior-Level Cleaning: This stage eliminates spurious interactions like accidental touches or idle playback. We apply a truncation strategy based on the 99.9th percentile of viewing duration, setting thresholds at 879s for videos and 9,601s for live streams. Outliers exceeding these limits are discarded, removing a median of 91 noisy records per user; (2) Text-Level Cleaning: Raw textual data, such as OCR, ASR, and captions, frequently contains inherent noise. To mitigate this, we adopt a cascade cleaning strategy. First, we apply regular expressions to eliminate structural noise, such as stuttering and garbled characters. Second, we utilize the Qwen2.5-72B-Instruct~\citep{yang2025qwen2} model to correct residual typos and semantic errors. This process achieves compression rates of 85.9\% for OCR text and 5.2\% for ASR text, significantly enhancing semantic density.

\noindent\textbf{Data Sampling.}
To balance fidelity and evaluation efficiency, we employ a clustering-based sampling strategy to select a diverse set of representative users as evaluation targets. We first encode each user to a comprehensive feature vector $v_u$ spanning four axes: 
(1) Demographics: User profiles including age and gender distributions; 
(2) Activity Level: Interaction intensity defined by total counts and active days; 
(3) Interest Distribution: Categorical preferences for content creators (e.g., Gaming, Playing, News); 
(4) Scenario Preference: Interaction frequency across the five major scenarios.
This multi-dimensional construction ensures coverage of diverse behavioral patterns.
Next, we apply K-Means clustering~\citep{mcqueen1967some} to the normalized vectors, partitioning users into behavioral clusters. The user nearest to each cluster centroid is selected as the representative, yielding a final set of 200 users.

\noindent\textbf{Data Anonymization.}
To safeguard user privacy and ensure ethical compliance, we implement a strict pipeline using a locally deployed Qwen3-235B \cite{yang2025qwen3} model.
First, we use it to identify sensitive entities (e.g., names, phone numbers, addresses) and replace them with semantic placeholders (e.g., \texttt{<NAME>}, \texttt{<PHONE>}, \texttt{<ADDRESS>}). Second, we assess content toxicity and automatically prune interactions flagged as "Harmful", such as hate speech and violence. Finally, we conduct manual verification to ensure strict adherence to ethical standards.

Based on the above processing, OmniBehavior encompasses the complete interaction trajectories of 200 users across 5 scenarios collected over a three-month period. Most notably, the dataset features an average sequence length of 8,143 actions, providing rich context for user behavior modeling.
Detailed distributions of sequence lengths and user demographics are visualized in Appendix \ref{appendix:exper_details} (Figures~\ref{fig:action_length_distribution_log} and~\ref{fig:demographic}), further underscoring the benchmark's long-horizon depth and population diversity.
\section{Authentic User Behavior Analysis}
To provide an in-depth understanding of authentic behavioral patterns, we conduct an empirical analysis of the OmniBehavior dataset. Specifically, we examine cross-scenario information gain and long-term causal dependencies. This analysis reveals that cross-scenario, long-horizon, and authentic data are essential for realistic user simulation.

\subsection{Multi-scenario Traces Enable Comprehensive User Modeling}
\label{sec:multi_scenario}

\begin{wrapfigure}[25]{r}{0.5\textwidth}
\centering
\vspace{-18pt}
\subfigure[User profile based on single-scenario vs multi-scenario.]{
    \begin{minipage}[b]{0.5\textwidth}
        \includegraphics[width=\linewidth]{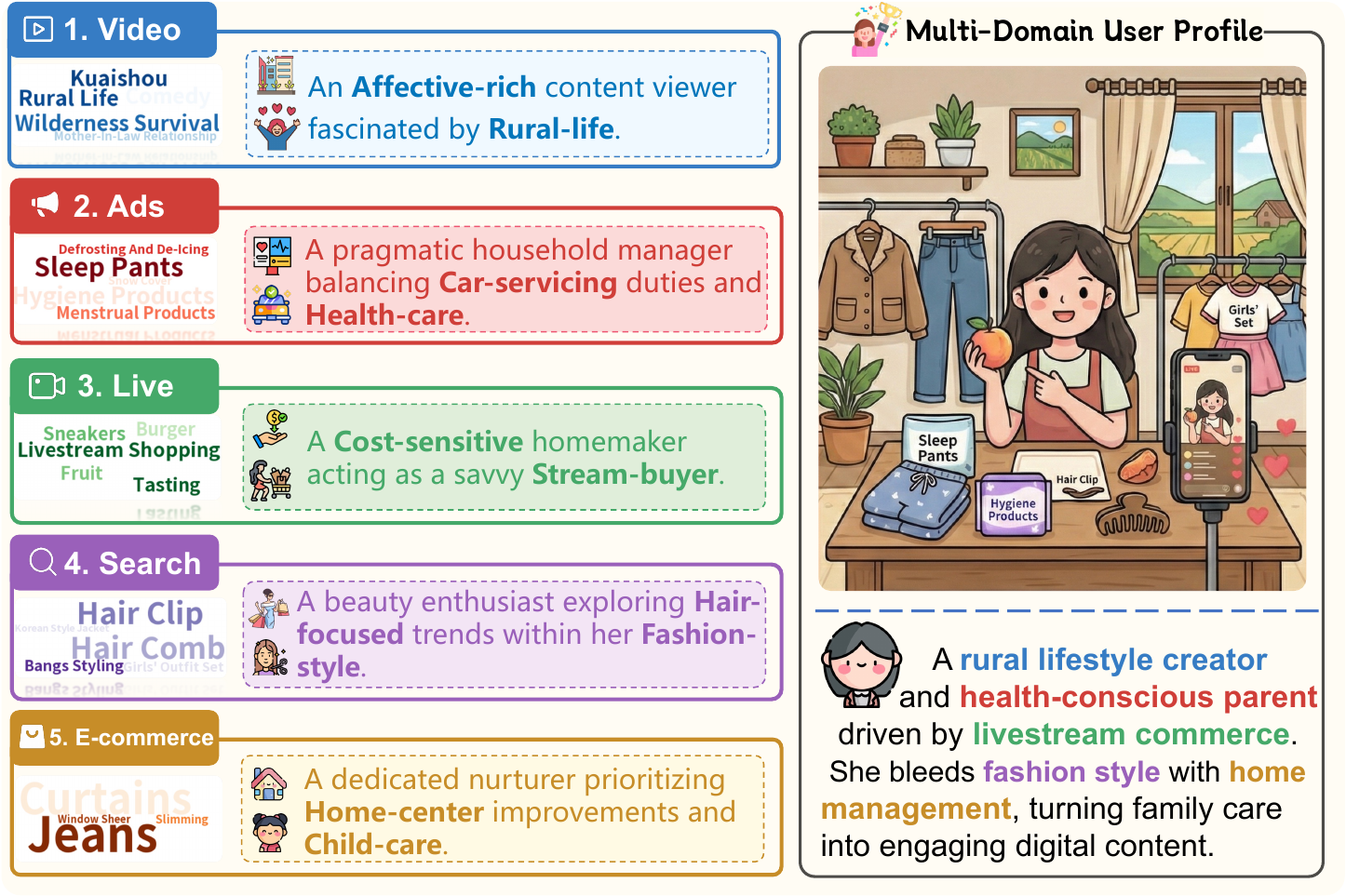}
    \end{minipage}
    \label{fig:multi_domain_a}
}

\subfigure[Cumulative interest coverage with increasing scenarios.]{
    \begin{minipage}[b]{0.5\textwidth}
        \includegraphics[width=\linewidth]{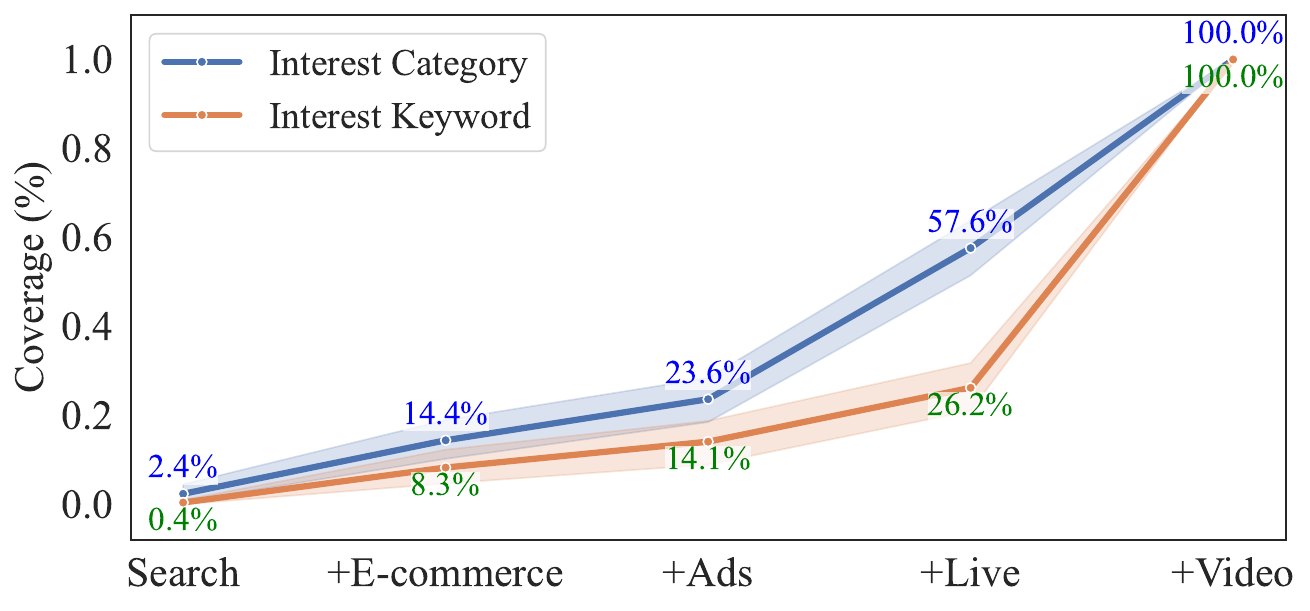}
    \end{minipage}
    \label{fig:multi_domain_b}
}
\vspace{-5pt}
\caption{Multi-scenario enables more comprehensive user modeling than single-scenario.}
\label{fig:case_study_multi_domain}
\end{wrapfigure}

To systematically assess whether single-scenario data is sufficient to model user preferences and behavioral patterns, we compare single-scenario and multi-scenario settings through qualitative profile reconstruction and quantitative measurements of interest coverage. Specifically, for each user, we extract one interest category and three keywords per interacted item using \texttt{Claude-Sonnet-4.5}, and aggregate them into user profiles.

As illustrated by the user profile example in Figure~\ref{fig:multi_domain_a}, profiles derived from single-scenario data are often fragmented and biased, whereas multi-scenario data provides richer contextual signals that better capture the user's stable and essential characteristics.

Furthermore, to examine whether this observation generalizes beyond individual cases, we perform a quantitative analysis. Specifically, we calculate the cumulative growth of unique interest categories and keywords as more scenarios are incorporated. The average results across all users are demonstrated in Figure~\ref{fig:multi_domain_b}, where information coverage increases consistently with scenario diversity, confirming the statistical significance of the multi-scenario advantage.

\begin{figure}[H]
  \centering
  \includegraphics[width=\textwidth]{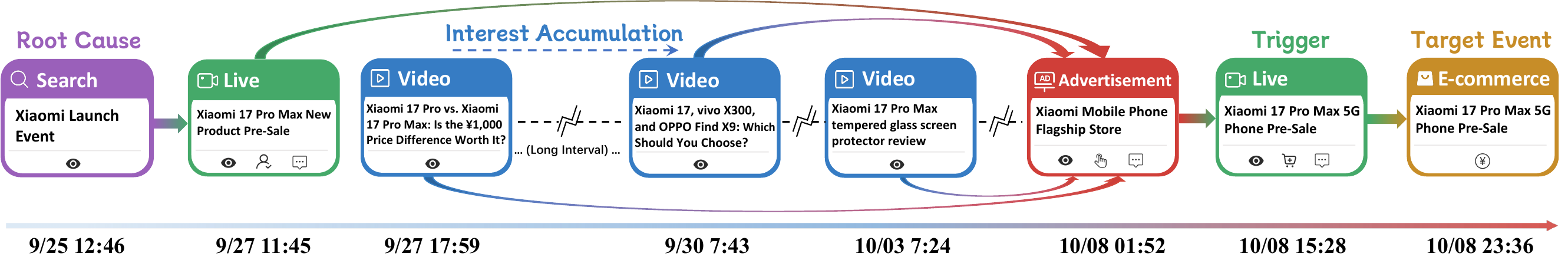}
  \caption{Case study of a cross-scenario causal chain, in which a search-initiated interest in “Xiaomi” accumulates across diverse interactions over 12 days and culminates in a purchase, highlighting the long-term and cross-scenario nature of user behavior.}
  \label{fig:xiaomi_causal_chain_analysis}
\end{figure}

\vspace{-2em}
\subsection{Causal Chains of User Behavior Span Across Scenarios and Long Sequences}
\label{sec:causal}

A profound understanding of user causal chains is a prerequisite for constructing a dataset that fully captures the authentic decision-making process. We randomly sample 180 high-value conversion events (e.g., ``Purchase'') and utilize \texttt{Claude-Sonnet-4.5} \cite{anthropic2025claudesonnet45} to identify the specific interaction paths within the full historical logs that lead to the event (which we term as causal chains). All traced paths are further manually verified to ensure correctness.

\begin{wrapfigure}[12]{r}{0.45\textwidth}
\vspace{-1em}
\centering
\includegraphics[width=\linewidth]{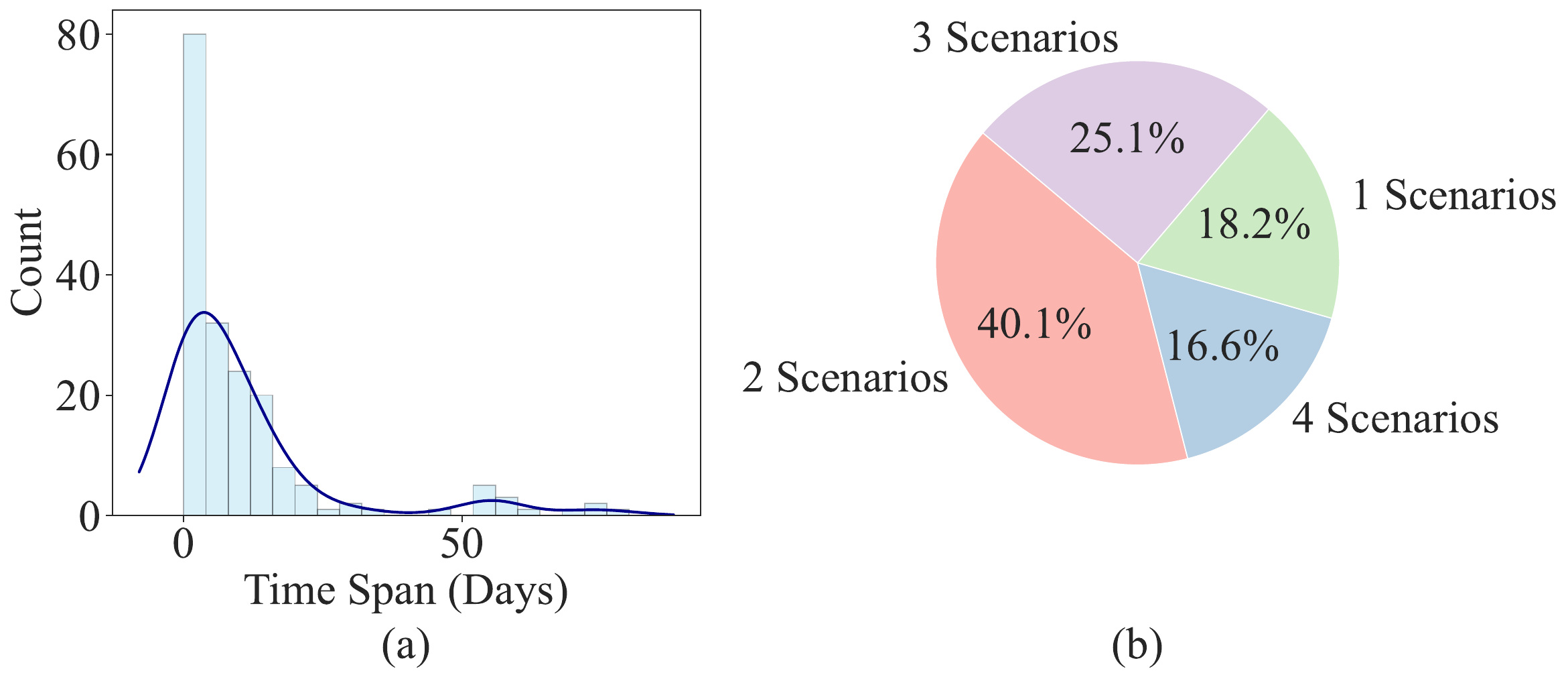}
\caption{Distributions of causal chain spans. (a) Time and (b) scenario counts demonstrate that user decisions naturally extend over long periods and across multiple scenarios.}
\label{fig:analysis_causal_chain}
\end{wrapfigure}

Figure~\ref{fig:analysis_causal_chain} illustrates the distributions of temporal spans and scenario spans of the extracted causal chains.
First, regarding temporal span, over 60\% of decisions rely on cues from more than 3 days prior, significantly exceeding the temporal window of traditional session-based assumptions \cite{pi2020search}. This necessitates a benchmark with sufficient temporal length. Second, cross-scenario dependency is the norm, with 81.8\% of chains spanning multiple scenarios. Together, these conclusions demonstrate that user decision-making is a naturally long-sequence and multi-scenario process, necessitating data that reflects this structure.

We further illustrate this with a representative case study in Figure~\ref{fig:xiaomi_causal_chain_analysis}, where a 12-day causal chain: spanning searches, cross-scenario interactions, and a live stream—culminates in a purchase. This confirms that decisions stem from long-term, cross-scenario accumulation, and that benchmarks limited to short sessions or single scenarios effectively perform "causal amputation" underscoring the necessity of ultra-long, multi-scenario environments.

\vspace{-0.2em}
\subsection{Distributional Divergence in Preference Evolution between Synthetic and Real User}

\label{appendix:diverage_of_synthetic_and_real_user}
\begin{wrapfigure}[13]{r}{0.56\textwidth}
\vspace{-1.3em}
\centering
\includegraphics[width=\linewidth]{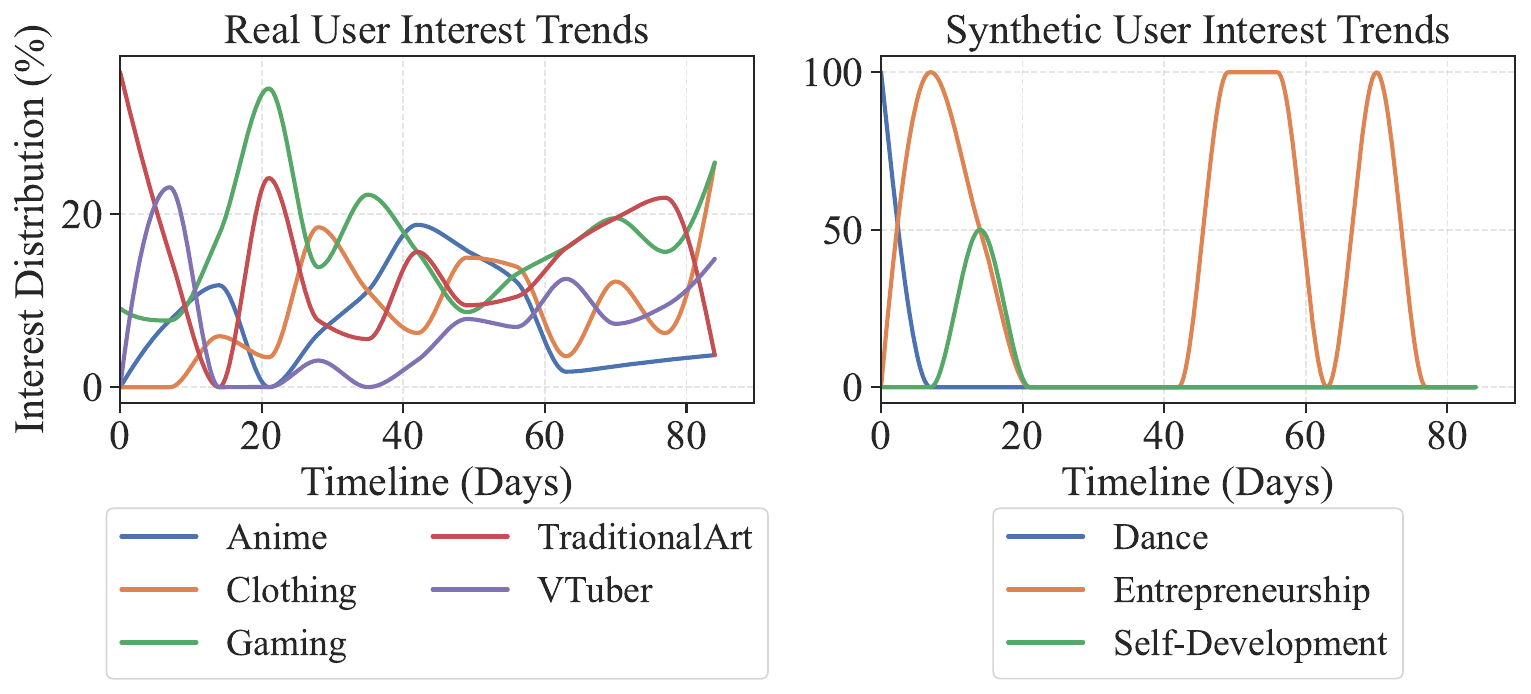}
\caption{Comparison of interest evolution in OmniBehavior (authentic, left) and LoCoMo (synthetic, right). Real preferences evolve smoothly, whereas synthetic interests appear as rigid, task-driven spikes.}
\label{fig:interest_drift}
\end{wrapfigure}

To investigate whether synthetic user simulation data can accurately reproduce the complex dynamics of real user interest evolution, we conduct a comparative analysis between authentic and synthetic user trajectories. Specifically, we compare OmniBehavior with LoCoMo~\citep{maharana2024evaluating}, a representative dataset that simulates long-horizon user memory through data synthesis. We adopt the interest extraction method described in Section~\ref{sec:multi_scenario} and track the temporal distribution of interest topics for both groups to characterize preference drift across the three-month observation period.

The resulting interest evolution patterns are shown in Figure~\ref{fig:interest_drift}. We extract interest categories from each interaction, map them into unified macro-level categories, and measure daily drift using $1-\text{Jaccard}$ between category sets on consecutive days. Real users exhibit smoother and more stochastic interest changes, with an average drift rate of 0.6311, while synthetic users show more rigid, abrupt shifts with a much lower drift rate of 0.1698. This gap suggests that synthetic users fail to capture realistic temporal interest dynamics, underscoring the need for grounding user simulators in real-world interaction data.

\section{LLM Evaluation}

\subsection{Experiment Settings}

\noindent\textbf{Task Definition.}
We formulate user simulation as predictive user simulation under an offline setting, where the simulator predicts a user's subsequent behavior conditioned on observable user information and the current scenario. Given a user profile \(p_u\), the user’s historical behavior sequence $\mathcal{H}_u$, and the context of a specific scenario \(c_t\), the simulator \(f_\theta\) aims to predict all corresponding user behavior(s) \(y_t\) in that scenario.
Formally, the historical behavior sequence is defined as
$\mathcal{H}_u=\{(c_1,a_1), \ldots, (c_{t-1},a_{t-1})\}$,
where each interaction consists of a scenario context $c_i$ and the realized user behavior $a_i$, covering both explicit signals (e.g., likes) and implicit signals (e.g., play time). The objective is to estimate
$\hat{y}_t = f_{\theta}(\mathcal{H}_u, c_t, p_u)$,
where $y_t$ may take the form of binary engagement decisions, continuous consumption durations, or textual feedback. 
Importantly, the user profile serves only as a weak prior alongside the historical behavior sequence and current-scenario context. We evaluate conditional predictability based on observable information. This offline formulation further enables controlled comparisons by exposing all models to the same observed histories and scenario contexts, avoiding additional variance introduced by environment feedback and platform policies.

\noindent\textbf{Metrics.}
We categorize user behaviors into 3 types, following prior settings~\citep{wang2025characterbox, wang2024rolellm}.
For binary behaviors (e.g., like, follow), we formulate simulation as a classification problem and report the F1-score.
For continuous behaviors (watch duration), standard absolute error introduces bias towards longer videos. To mitigate this, we report Normalized Mean Absolute Error (NMAE), normalizes the error by the video's duration.
Finally, for textual behaviors (e.g., customer service dialog), we utilize \texttt{Claude-Sonnet-4.5} as judge to quantify the performance of the simulator which measures four criteria (intent fidelity, persona mimicry, knowledge boundary, and semantic alignment, in Figure~\ref{fig:user_sim_judge}).

\begin{table*}[t]
\captionsetup{belowskip=0pt, aboveskip=6pt}
\caption{Comparison between 11 LLMs and humans on the OmniBehavior benchmark. We categorize user behaviors into three types: binary behaviors (e.g., clicks), continuous behaviors (e.g., duration), and textual behaviors (e.g., dialogue). The overall score represents the aggregated performance. Human$^\ast$ denotes the average performance of three human annotators on 100 sampled tasks from each scenario. The best/second best scores are \textbf{bolded}/\underline{underlined}.}
\setlength{\tabcolsep}{9pt}
\setlength{\extrarowheight}{1pt}
\setlength{\aboverulesep}{0pt}
\setlength{\belowrulesep}{0pt}
\resizebox{\textwidth}{!}{
\begin{tabular}{l|cc|c|c|cc|c}
\toprule
 \multirow{2}{*}{Model} & \multicolumn{2}{c|}{Video} & Live & Ads & \multicolumn{2}{c|}{E-commerce} & \\
 & Binary & Continuous & Binary & Binary & Binary & Textual & \multirow{-2}{*}{Overall Score} \\
\toprule
\rowcolor{black!7}\multicolumn{8}{l}{\emph{\underline{Human baseline}}} \\
Human$^*$ & 50.21 & 68.83 & 46.79 & 54.76 & 41.46 & 74.63 & 56.11 \\
\rowcolor{black!7}\multicolumn{8}{l}{\emph{\underline{Closed-source}}} \\
Claude-Opus-4.5 & \textbf{33.05} & 64.19 & \textbf{31.70} & \textbf{51.16} & 29.98 & \textbf{57.21} & \textbf{44.55} \\
Claude-Sonnet-4.5 & 18.85 & \textbf{65.95} & 25.00 & \underline{42.77} & \textbf{36.13} & 54.26 & 40.49 \\
Claude-Haiku-4.5 & 22.84 & 63.26 & 26.11 & 30.00 & 26.37 & 50.29 & 36.48 \\
Claude-Sonnet-4 & 25.29 & 64.62 & 28.86 & 36.81 & 16.50 & 49.13 & 36.87 \\
Gemini-3-Flash & 22.09 & 53.79 & 25.61 & 24.64 & 19.65 & 49.80 & 32.60 \\
GPT-5.2 & \underline{31.54} & \underline{65.01} & 28.63 & 33.60 & 29.32 & 46.29 & 39.07 \\
GPT-4o & 27.88 & 62.75 & 28.15 & 25.24 & 28.66 & 44.92 & 36.27 \\

\rowcolor{black!7}\multicolumn{8}{l}{\emph{\underline{Open-source}}} \\
GLM-4.7 & 26.86 & 64.43 & \underline{28.97} & 40.34 & 32.90 & \underline{55.25} & \underline{41.46} \\
DeepSeek-V3 & 21.42 & 63.98 & 27.92 & 25.74 & \underline{33.31} & 52.13 & 37.42 \\
Kimi-K2-Instruct & 23.30 & 64.80 & 28.60 & 31.19 & 29.94 & 47.83 & 37.61 \\
Qwen3-235B & 18.26 & 62.38 & 23.84 & 23.19 & 19.22 & 45.74 & 32.11  \\
\bottomrule
\end{tabular}}
\label{tab:result_all}
\end{table*}

\noindent\textbf{Baselines.}
We evaluate a diverse set of representative LLMs, including both closed-source and open-source models. The closed-source models include the Claude series (Claude-Opus-4.5~\citep{anthropic2025claudeopus45}, Claude-Sonnet-4.5~\citep{anthropic2025claudesonnet45}, Claude-Haiku-4.5~\citep{anthropic2025claudehaiku45}, and Claude-Sonnet-4~\citep{anthropic2025claude4}), as well as Gemini-3-Flash~\citep{google2025gemini3flash} and the GPT series (GPT-5.2~\citep{gpt5} and GPT-4o~\citep{achiam2023gpt,hurst2024gpt}). The open-source models include GLM-4.7~\citep{zai2025glm47}, DeepSeek-V3~\citep{liu2024deepseek}, Kimi-K2-Instruct-0905 (hereafter Kimi-K2-Instruct)~\citep{team2025kimi}, and Qwen3-235B-A22B-Instruct-2507 (hereafter Qwen3-235B)~\citep{yang2025qwen3}. We additionally include a human baseline to contextualize model performance. Specifically, we recruit three annotators with master's degrees in engineering and sample 100 tasks from each scenario. Since each input can contain up to 32K tokens, we present the information through a structured and searchable Gradio interface. We report the average score of the three annotators as the human baseline. More details are provided in Appendix~\ref{appendix:case_study}.

\noindent\textbf{Implementation Details.}
We evaluate models on 6,000 behavior prediction tasks across different scenarios. To ensure the robustness of our evaluation, we carefully sample these tasks by enforcing temporal balance, scenario balance, and value distribution balance. We adopt a global 32k context window for main experimental evaluation and implement the memory management using the LangChain framework~\citep{mavroudis2024langchain}. All experiments are conducted on NVIDIA A800-SXM4-80GB machines. Prompts and hyperparameter settings used in the experiments are provided in Appendix~\ref{appendix:experimental_details}.

\subsection{Overall Results}

Table \ref{tab:result_all} reports the overall performance of diverse LLM backbones on OmniBehavior. Results show that high-fidelity user simulation remains highly challenging. Even the best-performing model, Claude-Opus-4.5 achieves an overall score of 44.55, with most models clustered between 32 and 41. This low range indicates that general instruction tuning is insufficient for modeling the stochasticity, long-tail distributions, and cross-scenario causal dependencies inherent in human behavior, revealing a substantial gap between current LLMs and real human patterns.

The human baseline further demonstrates a substantial gap between current LLMs and human performance under the same observational conditions. Notably, this human performance should not be interpreted as the fundamental upper bound of the prediction task, but rather as an empirical upper bound under the current evaluation setting. Our case studies suggest that human errors mainly arise from difficulties in retrieving historical evidence and integrating fine-grained individual-level signals distributed across long-horizon behavior sequences (see Appendix \ref{appendix:human_errors} for representative cases).

Closed-source models set the overall performance ceiling, with Claude-Opus-4.5 achieving the highest score of 44.55 and leading 4 of 6 metrics. Yet this advantage is not absolute: the open-source GLM-4.7 ranks second overall with 41.46, outperforming Claude-Sonnet-4.5 (40.49) and GPT-5.2 (39.07). Open-weight models also excel in targeted settings, as DeepSeek-V3 surpasses Claude-Opus-4.5 in E-commerce binary behavior simulation (33.31 vs. 29.98). Thus, while closed-source models dominate overall, optimized open-weight models can rival or exceed them on specific tasks. Looking forward, with continued advances in long-context modeling, external memory mechanisms, and individual-level behavior modeling, we believe that further approaching or even surpassing current human annotation performance represents a plausible research direction.

\subsection{Evaluation of Long Context Modeling}

\begin{wrapfigure}[12]{r}{0.48\textwidth}
\vspace{-1em}
\centering
\includegraphics[width=\linewidth]{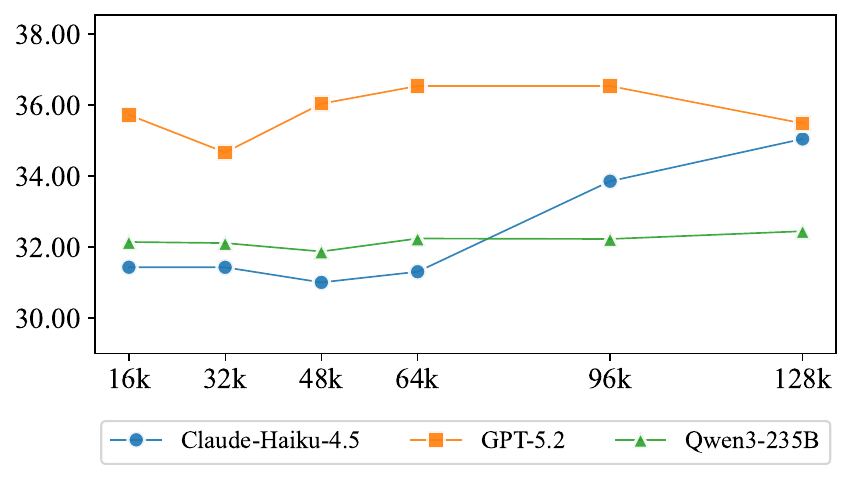}
\caption{Effect of context size.}
\label{fig:effect_len}
\end{wrapfigure}

\paragraph{Effect of Context Window Size.}
We evaluate representative open-source and closed-source LLMs with context windows from 16K to 128K tokens on a subset of users ($N=66$) whose interaction histories exceed 128K tokens. As shown in Figure~\ref{fig:effect_len}, longer contexts do not consistently improve performance, suggesting that current LLMs, despite supporting ultra-long inputs, still struggle to effectively leverage extended interaction histories for long-context reasoning, consistent with prior findings~\citep{liu2024lost}. Appendix~\ref{appendix:context_case_study} further illustrates how models use newly introduced long-range evidence as context expands.

\begin{wrapfigure}[10]{r}{0.48\textwidth}
\vspace{-1.4em}
\centering
\includegraphics[width=\linewidth]{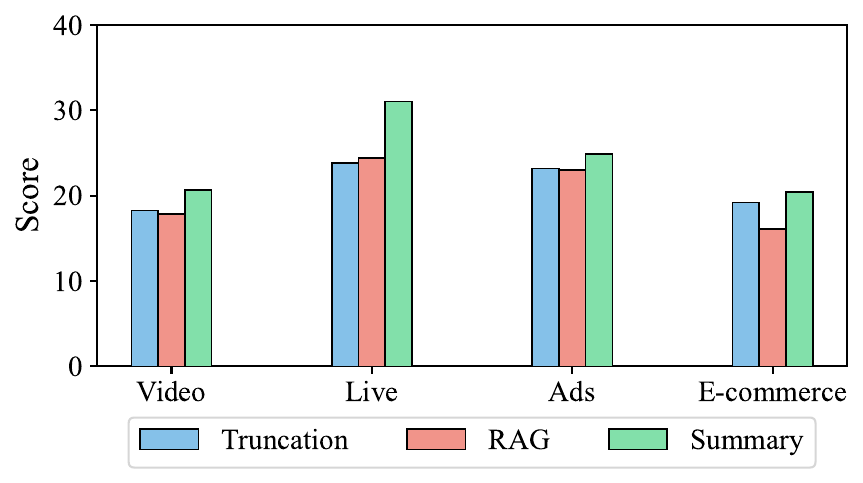}
\vspace{-1.6em}
\caption{Effect of memory management.}
\label{fig:memory_management}
\end{wrapfigure}

\paragraph{Memory Management Mechanisms.}
Based on Qwen3-235B, we compare two memory management approaches: history summarization and RAG-based retrieval of relevant scenarios, with implementation details in Appendix~\ref{appendix:memory_management_app}. Figure~\ref{fig:memory_management} shows that both offer only modest gains, as RAG relies on semantic similarity and ignores causal dependencies, while summarization may discard critical behavioral details. These results highlight the need for mechanisms that explicitly preserve the causal and temporal structure of user history.

\subsection{Historical Information Ablation}
\begin{wraptable}[8]{r}{0.3\textwidth}
\vspace{-3.3em}
\centering
\caption{Effect of historical behavioral information.}
\vspace{-0.5em}
\footnotesize
\setlength{\tabcolsep}{2pt}
\begin{tabular}{lccc}
\toprule
& \textbf{\shortstack{Own\\History}}
& \textbf{\shortstack{No\\History}}
& \textbf{\shortstack{Other\\History}} \\
\midrule
\multicolumn{2}{l}{Qwen3-235B} & & \\
\hdashline
AUC & \textbf{0.67} & 0.52 & 0.49 \\
F1  & \textbf{0.26} & 0.12 & 0.12 \\
\midrule
\multicolumn{2}{l}{Qwen3-8B} & & \\
\hdashline
AUC & \textbf{0.69} & 0.53 & 0.52 \\
F1  & \textbf{0.23} & 0.10 & 0.09 \\
\bottomrule
\end{tabular}
\label{tab:history_ablation}
\end{wraptable}
To assess whether historical behavior provides individualized signals beyond demographic profiles and current-scenario information, we compare the full setting with two ablations: removing history (No History) and replacing it with another user's history (Other History). As shown in Table \ref{tab:history_ablation}, both consistently degrade performance across models and metrics, indicating that users' own behavioral histories provide substantial predictive signals beyond group-level priors.

\section{The Structural Bias of LLM Simulator}
To probe the underlying causes of the simulation gap, we conduct a systematic comparison between simulated and authentic behaviors, uncovering a fundamental structural bias: a "\textit{positivity-and-average}" tendency in current LLM-based simulators. We demonstrate this bias by comparing behavior probability and emotional distributions between simulated and real users, and quantifying LLMs’ ability to model behavioral differences across users.

\paragraph{Hyper-activity Bias.}

\begin{wrapfigure}[9]{r}{0.45\textwidth}
\vspace{-1.5em}
\centering
\includegraphics[width=\linewidth]{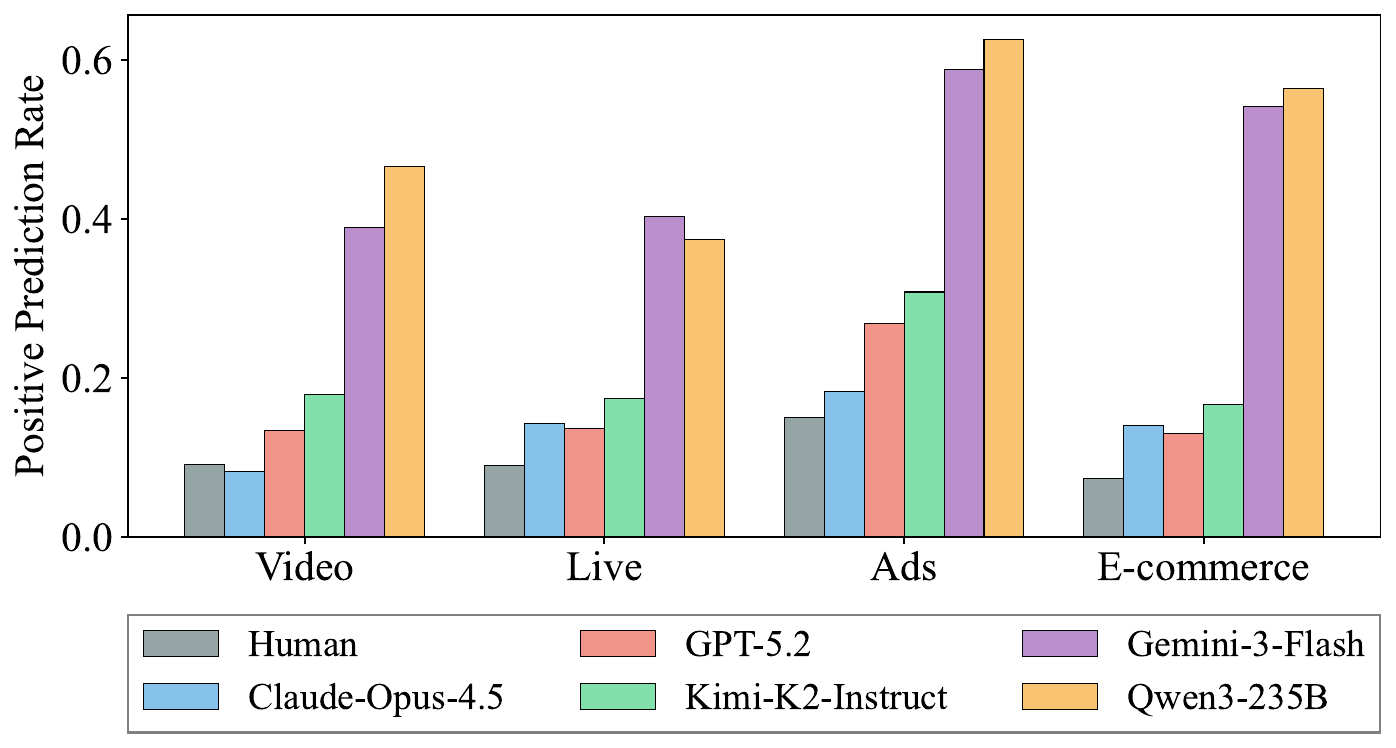}
\vspace{-1.5em}
\caption{LLM simulators overestimate positive user interactions across scenes.}
\label{fig:overestimation}
\end{wrapfigure}

We compare real and simulated behavior distributions via positive interaction rates (positive outcomes among all behavior instances). Figure~\ref{fig:overestimation} shows that real behavior is sparse, with positive rates below 10\%, while LLM simulators such as Qwen3-235B and Gemini-3-Flash overestimate user engagement by 40--60\%. This hyper-activity bias shows poor modeling of implicit rejection, limiting their usage like user churn warning~\citep{sharma2023towards}.

\paragraph{Utopian Tendency.}

\begin{wrapfigure}[12]{r}{0.45\textwidth}
\vspace{-0.5em}
\centering
\includegraphics[width=\linewidth]{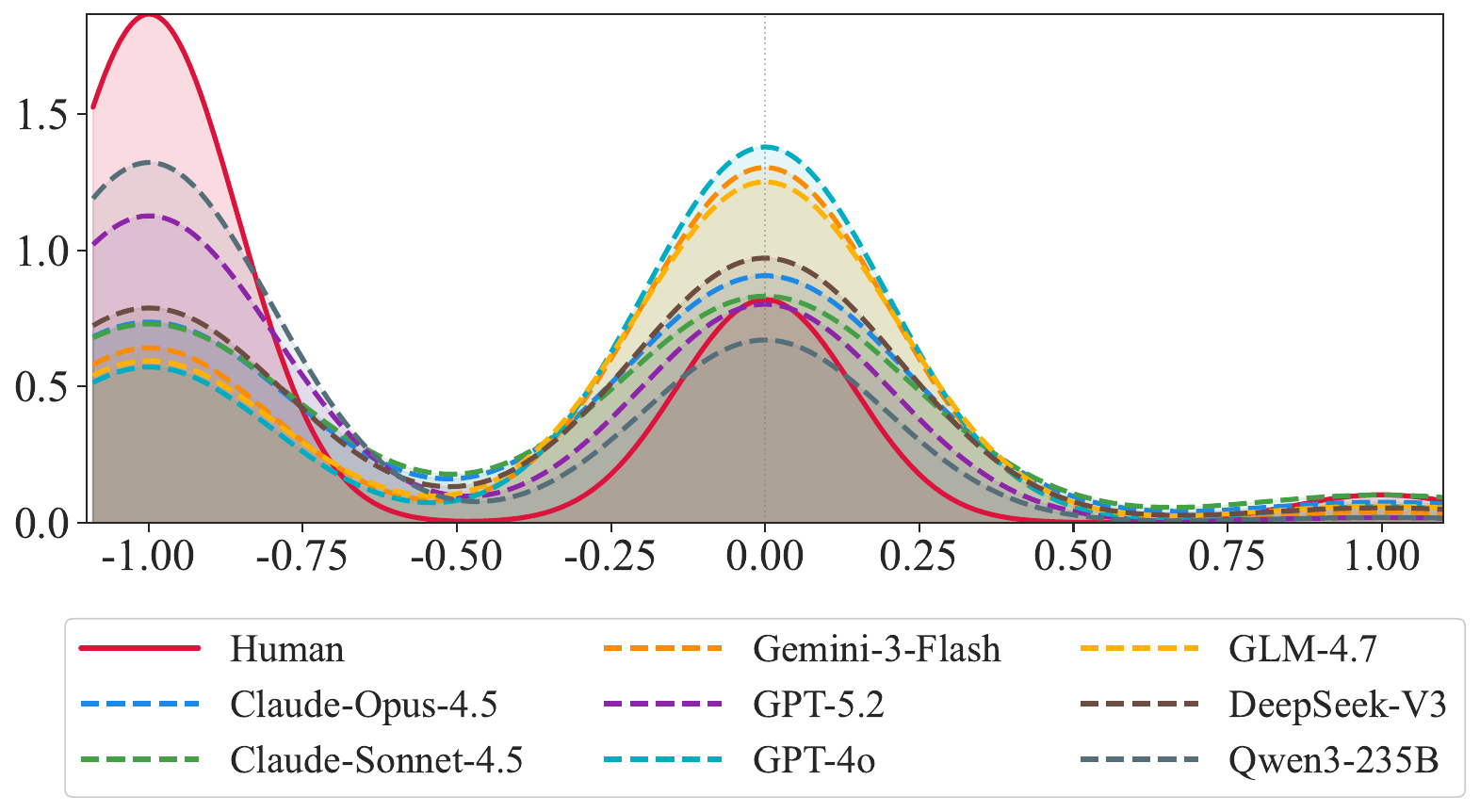}
\vspace{-1.5em}
\caption{LLM dialogues are more positive and miss real users' negative emotions.}
\label{fig:sentiment}
\end{wrapfigure}

We further examine affective realism in E-commerce customer service dialogues by analyzing emotional expression and language style. Using \texttt{Claude-Sonnet-4.5}, we assign each utterance a sentiment label in \{-1,0,1\}, from negative to positive. Figure~\ref{fig:sentiment} shows that real users often express strong negative emotions, while LLM-generated utterances cluster around neutral or positive sentiment. Overall, language style analysis further shows that LLM outputs are more polite, whereas real users are more direct and emotionally expressive (Appendix \ref{appendix:utopian_tendency_results}), producing an overly sanitized environment for adversarial social dynamics.

\paragraph{Personality Homogenization.}

\begin{wrapfigure}[14]{r}{0.45\textwidth}
\vspace{-1.5em}
\centering
\includegraphics[width=\linewidth]{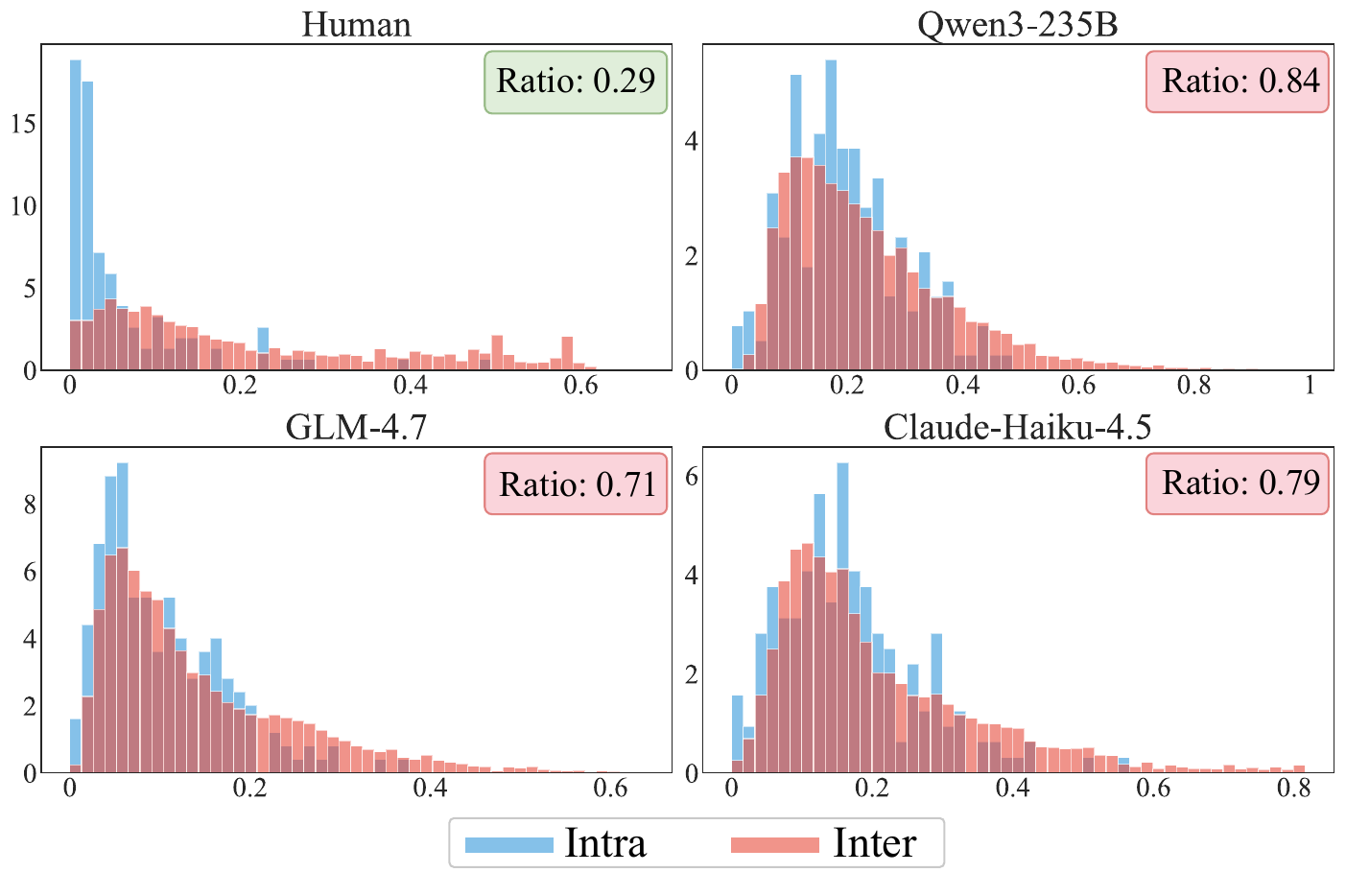}
\vspace{-1.8em}
\caption{Comparison of Intra-user and Inter-user behavioral distances for real users and LLM-simulated users.}
\label{fig:intra_inter_multi_model_comparison}
\end{wrapfigure}

These findings raise whether LLM simulators preserve personalized user characteristics. We analyze intra- and inter-user behavioral distances using 19-dimensional action rate vectors (e.g., like, share, activate). Intra-user distance measures consistency within a user's history, while inter-user distance captures cross-user heterogeneity. Figure~\ref{fig:intra_inter_multi_model_comparison} shows that real users have much larger inter-user than intra-user variation (Inter $\gg$ Intra, Ratio $\approx$ 0.29), whereas LLM-generated users show heavily overlapping distributions (Ratio $\approx$ 0.7$-$0.87). This indicates homogenized simulations that fail to preserve distinct user identities, likely due to generic pre-training patterns suppressing long-tail personalized signals. Results of other models are provided in Appendix~\ref{appendix:personality_results}, demonstrating the same trends.

\section{Related Works}
\vspace{-0.1em}

\subsection{LLMs as Human Simulators}
Advancements in LLMs have enabled agents to simulate human cognition and interaction across diverse domains, including dialogue interactions~\citep{chan2023chateval,zhang2026ai}, recommender systems~\citep{wang2023large, zhang2024agentcf}, autonomous driving~\citep{jin2023surrealdriver}, user understanding~\citep{tan2025personabench}. Seminal works like Generative Agents~\citep{park2023generative} and BASES~\citep{ren2024bases} further extend this to social behaviors and web search. 
However, these approaches primarily operate within controlled, single-task environments or synthetic sandboxes. They often fail to capture the stochasticity, long-tail distributions, and cross-scenario dependencies inherent in real-world digital footprints. 
This gap underscores the need for benchmarks grounded in authentic industrial logs to rigorously assess alignment with real human behaviors. Our work addresses this need by providing a cross-scenario, real-world benchmark designed to evaluate the capabilities and limitations of LLMs in capturing diverse and long-term human behavior patterns.

\subsection{Evaluation for Human Simulation}
Existing evaluations fall into two categories: log-based behavioral prediction and role play benchmarks.
The first category evaluates action prediction using real-world logs. These benchmarks have evolved from static interaction matrices (e.g., MovieLens~\citep{harper2015movielens}, Netflix~\citep{bennett2007netflix}) to sequential logs (e.g., Tenrec~\citep{yuan2022tenrec}, Alibaba-CCP~\citep{zhou2018deep}) and interactive environments (e.g., VirtualTaobao~\citep{shi2019virtual}, RecSim~\citep{ie2019recsim}, Sim4IA-Bench~\citep{kruff2025sim4ia}). Domain-specific evaluations have also extended to conversational recommendation~\citep{yoon2024evaluating} and social movement simulation on Twitter~\citep{mou2402unveiling}.
However, a critical limitation is their confinement to isolated scenarios. By fragmenting user behaviors into disconnected silos (e.g., solely e-commerce), they miss cross-scenario causal dependencies—such as a purchase triggered by a short video—thereby failing to support holistic behavior modeling.
Role-playing benchmarks, in contrast, focus on persona consistency, memory, and multi-turn interaction~\citep{xie2024human,maharana2024evaluating,chen2025consistentchat,hu2025simbench}, but often rely on synthesized profiles or standardized tasks rather than noisy longitudinal logs.
OmniBehavior bridges these gaps with large-scale real-world logs that capture cross-scenario, long-horizon, and heterogeneous behavioral patterns, enabling more ecologically valid evaluation of LLM simulators.

\vspace{-0.5em}
\section{Discussion}
\label{sec:discussion}
\vspace{-0.5em}

\noindent\textbf{Conclusions.} We present \textbf{OmniBehavior}, a real-world benchmark for evaluating predictive user simulation. 
Our statistical analysis shows that human behavior exhibits long-term and cross-scenario dependencies, suggesting that single-scenario settings and purely synthetic data are insufficient for realistic application. Experimental results reveal that current LLMs exhibit a substantial capability gap in modeling real-world user behaviors, regardless of context length. 
Furthermore, we identify structural bias in LLM-based simulators, characterized by a "positivity-and-average" tendency that homogenizes users and overestimates engagement. 
Such biases limit the reliability of LLM-based simulators for real-world scenarios. We hope OmniBehavior will encourage future work on modeling the long-horizon, cross-scenario, and heterogeneous nature of human behavior.

\noindent \textbf{Limitations.} Due to the rapid evolution of the field, the models and memory management algorithms covered in our current evaluation do not yet encompass all state-of-the-art approaches. To address this limitation, we plan to maintain a public leaderboard that supports community-contributed evaluation submissions, enabling continuous tracking and timely updates of the latest methods.
In addition, OmniBehavior is based on real user behavior from Kuaishou and cannot fully represent diverse cultural and linguistic contexts. Extending the benchmark to other countries, languages and platforms is an important direction~\citep{tang2019towards,qin2026restoring,chopra2026beyond,ivey2026real,zhao2025comparing}.
Finally, OmniBehavior adopts an offline predictive formulation that does not capture closed-loop interactions. Extending it to interactive evaluation while preserving privacy and reproducibility remains future work~\citep{jin2025beyond,barres2026taubench,chen2026appworldul}.

\section*{Acknowledgements}
We sincerely thank the reviewers for their insightful comments and valuable suggestions. This work was supported by the Natural Science Foundation of China (No. 62536008, 62476265, 62306303), the Postdoctoral Fellowship Program of CPSF under Grant Number GZB20260380, Beijing Nova Program No. 202604841270, the Basic Research Program of ISCAS, Grant No. ISCAS-JCZD-202602.

{
\bibliographystyle{plainnat}
\bibliography{ref}
}
\newtcolorbox{AIBox}[2][]{aibox,title=#2,#1}
\tcbset{
  aibox/.style={
    width=0.95\textwidth,
    top=5pt,
    colback=black!05,
    colframe=black!20,
    colbacktitle=black!50,
    enhanced,
    center,
    attach boxed title to top left={yshift=-0.1in,xshift=0.1in},
    boxed title style={boxrule=0pt,colframe=white,},
  }
}

\appendix
\section*{LLM Usage Statement}
\label{sec:llm_usage}
Large Language Models were used only for limited auxiliary purposes during the preparation of this manuscript. In particular, the authors used LLMs to assist with language editing, including grammar correction, spelling checks, as well as to provide coding assistance during the experimental implementation process. All generated text and code suggestions were carefully reviewed, revised, tested, and approved by the authors. The LLMs did not contribute to the formulation of the research questions, conceptual ideas, experimental design, interpretation of results, or scientific conclusions. All research-level contributions, including the ideas, methodology, analysis, and final decisions, were made by the whole authors.

\section*{Ethics Statement}

This work is based on real user behavior data collected and processed under applicable legal, platform, and privacy-protection requirements. The dataset was reviewed by the internal legal and data compliance teams, and its use for research and benchmark construction was confirmed to be consistent with applicable local laws and the Kuaishou User Agreement and Privacy Policy.

The data underwent multiple stages of anonymization, privacy-preserving cleaning, and manual review (more details are provided in Section \ref{sec:construction_pipeline}). Personally identifiable and sensitive information, including names, phone numbers, addresses, device information, and other user-specific identifiers, was removed or replaced with semantic placeholders. Five legal professionals further conducted record-level review of the evaluation data, followed by multiple rounds of revision and re-review. No residual personally identifiable information was identified in the final processed data. The anonymized dataset was also formally reviewed and approved by the institutional ethics committee and completed the required internal compliance and release procedures.
The final dataset was assessed to substantially mitigate re-identification risks through irreversible removal of direct and externally linkable identifiers and manual review of sensitive textual content. All experiments were conducted exclusively on the processed data. 

\textbf{Use of this dataset indicates your agreement to comply with all applicable privacy, data protection, and research ethics requirements.}

\clearpage
\section{Data Statistics}
\label{appendix:exper_details}

\subsection{Action Sequence Length Distribution}

\begin{wrapfigure}[18]{r}{0.48\textwidth}
\centering
\includegraphics[width=\linewidth]{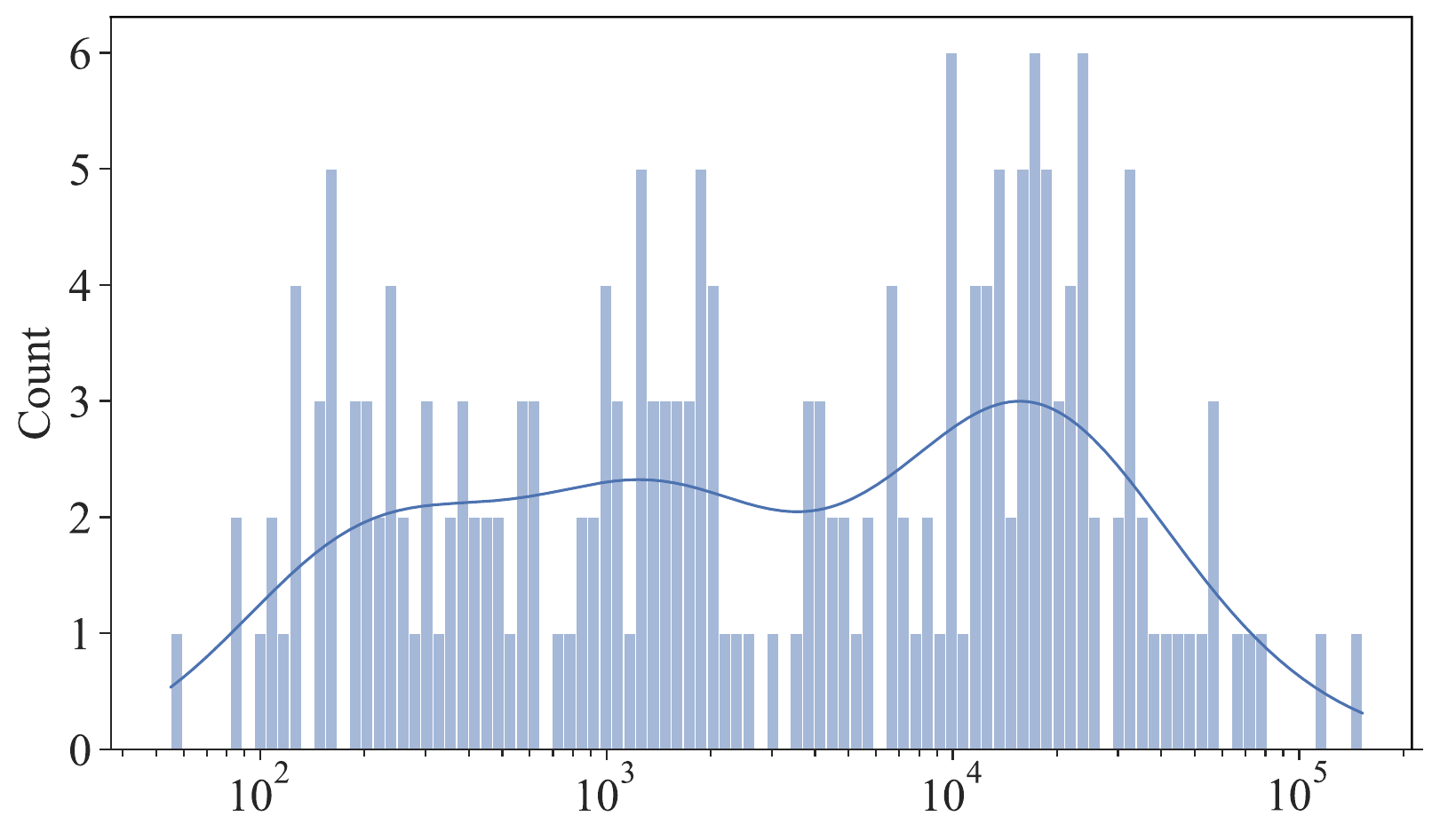}
\caption{Log-scaled distribution of user action sequence lengths, spanning hundreds to over 100k, requiring models to handle ultra-long contexts.}
\label{fig:action_length_distribution_log}
\end{wrapfigure}

To provide a clearer picture of OmniBehavior's composition, we report additional statistics on user behavior sequences and population attributes. Figure~\ref{fig:action_length_distribution_log} illustrates that user action sequence lengths span multiple orders of magnitude on a logarithmic scale, range from relatively short traces to extremely long histories that can exceed 100k steps.
Such characteristics reflect the inherently long-horizon and heterogeneous nature of real-world user behavior. They also underscore the necessity for models to effectively capture both short-term dynamics and long-range dependencies within a unified framework. Consequently, OmniBehavior presents substantial challenges for long-context modeling, particularly in terms of memory efficiency, temporal reasoning, and the ability to maintain coherent representations over ultra-long sequences.

\subsection{User Demographics and Interest Diversity}
\begin{figure}[H]
  \centering
  \includegraphics[width=0.9\textwidth]{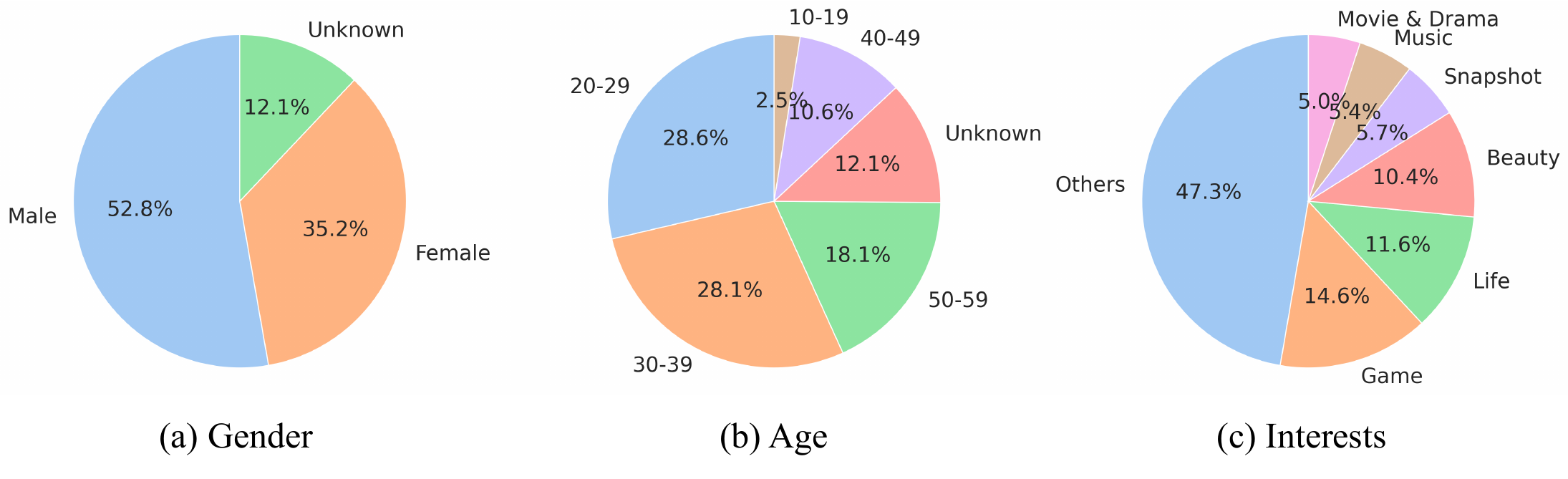}
  \caption{Demographic and behavioral distributions of users in the benchmark. The charts show diversity across gender, age groups, interests, ensuring a representative population for user simulation.}
  \label{fig:demographic}
\end{figure}

Figure~\ref{fig:demographic} complements this view by presenting the demographic and interest distributions of the sampled users. The results show that OmniBehavior covers users from diverse gender and age groups, while also spanning a broad spectrum of content interests. Such diversity shows that the dataset does not concentrate on a narrow user segment, but instead preserves substantial heterogeneity in both user background and preference structure. Taken together, these statistics indicate that our OmniBehavior benchmark is challenging not only because of its long behavioral horizons, but also because it captures a diverse and heterogeneous user population, which is essential for realistic user behavior simulation.

\section{Benchmark Schema}
\label{appendix:benchmark_schema}

\begin{figure}[H]
\centering
\includegraphics[width=0.9\textwidth]{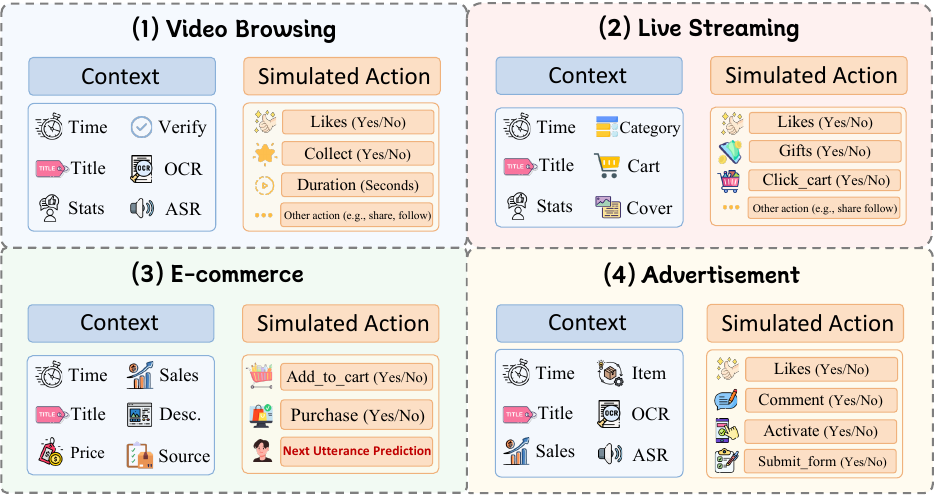}
\caption{The OmniBehavior Benchmark Scope. We construct a unified predictive user simulation benchmark covering multiple major user activities on the Kuaishou platform, with customer service next-utterance prediction treated as part of the E-commerce scenario. The framework requires LLMs to predict diverse behaviors (e.g., watch duration, purchase, comment, and customer-service responses) based on scenario-specific contexts, serving as a comprehensive testbed for high-fidelity user simulation.}
\label{fig:case_all_domain}
\end{figure}

Figure~\ref{fig:case_all_domain} illustrates the detailed schema of the OmniBehavior benchmark across representative user scenarios on the Kuaishou platform, including Video Browsing, Live Streaming, E-commerce, and Advertisement. Search Behavior is not explicitly visualized because it does not involve a rich scenario context beyond the search query itself. Customer service next-utterance prediction is treated as part of the E-commerce scenario.

As shown in the figure, the input context consists of heterogeneous signals describing the interaction environment, such as timestamps, content titles, item attributes, and historical statistics, along with textual signals extracted from multimodal sources (e.g., OCR and ASR transcripts when available). Based on these contextual signals, the LLM-based simulator is required to generate simulated user actions. The action space varies across scenarios and includes behaviors such as liking, collecting, commenting, purchasing, and predicting watch duration or conversational responses.

This unified formulation enables OmniBehavior to capture diverse interaction patterns across multiple product surfaces while maintaining a consistent evaluation protocol. The resulting benchmark provides a structured and realistic environment for assessing the capability of LLM-based user simulators in modeling complex user behaviors.

\vspace{-2em}
\section{Experimental Parameters and Details}
\label{appendix:experimental_details}

\subsection{Model Inference Details}
In all main experimental evaluations, model inference was conducted via API calls with a fixed temperature of 0.1, while the top-p parameter was kept at each model’s default setting to ensure consistency and reduce sampling variance. Closed-source models, as well as DeepSeek-V3 and Kimi-K2-Instruct, were accessed through their official public APIs. In contrast, GLM-4.7 and Qwen3-235B were deployed locally using the vLLM framework \citep{kwon2023efficient} on NVIDIA A800-SXM4-80GB machines, enabling efficient large-scale inference under controlled hardware conditions.

\subsection{Memory Management Results}
\label{appendix:memory_management_app}
We evaluate three representative memory management strategies: \textbf{(1) Truncation}, retains only the most recent context within the fixed window; \textbf{(2) RAG}, retrieves top-$k$ relevant interactions ($k$=100) to ensure high recall over long user histories; and \textbf{(3) Summary}, incrementally compresses user history via periodic summarization with a 4K token buffer. For the RAG module, dense retrieval is implemented using the bge-m3 embedding model (1,024 dimension). The retrieval stage returns Top-100 candidates with additional oversampling to improve recall, while vector indices are cached and periodically refreshed to balance efficiency. For the summarization module, we employ Qwen3-32B (no-thinking) with the same temperature setting to ensure stable outputs. A buffer-based incremental strategy is adopted, where summarization is triggered once the buffer reaches its 4K-token limit, and each summary is constrained to at most 1,000 tokens.

As shown in Table~\ref{tab:result_rag_summary}, different memory management strategies exhibit distinct trade-offs across scenarios. All methods are implemented using the same model (Qwen3-235B) in the main experiment and evaluated on the full user interaction history to ensure fairness. The Summary method achieves the best overall performance, particularly yielding substantial improvements in the Live Streaming scenario (31.06 vs. 23.84), suggesting its potential advantage in aggregating long-horizon interaction signals. In contrast, RAG provides marginal or even negative gains compared to Truncation in certain scenarios (e.g., E-commerce, 16.14 vs. 19.22), which may stem from retrieval noise or the disruption of sequential behavioral dependencies inherent in dense-retrieval-based context construction.

Overall, the results suggest that existing memory management strategies are insufficient for modeling real-world user behavior. Truncation preserves fidelity but lacks long-term reasoning, RAG retrieves relevant fragments but ignores causal structure, and summarization captures global trends but inevitably loses fine-grained behavioral details. 
These limitations call for the need of structure-aware memory mechanisms that explicitly model temporal order, cross-scenario dependencies, and causal relationships in user behavior sequences. We will explore more advanced memory architectures to improve long-horizon modeling in the future work.

\begin{table}[htbp]
\captionsetup{belowskip=0pt, aboveskip=6pt}
\caption{Comparison of memory management strategies on OmniBehavior using Qwen3-235B. Each scenario is formulated as a binary behavior prediction task (e.g., like, collect, share).}
\centering
\resizebox{0.8\textwidth}{!}{
\begin{tabular}{lccccc}
\toprule
Method & Video & Live & Ads & E-commerce & Avg. \\
\midrule
Truncation & 18.26 & 23.84 & 23.19 & 19.22 & 21.13  \\
\hdashline
RAG & 17.9 & 24.44 & 23.03 & 16.14  & 20.38 ($\downarrow$ 3.60\%) \\
Summary & 20.7 & 31.06 & 24.86 & 20.45  & 24.27 ($\uparrow$ 14.9\%)\\
\bottomrule
\end{tabular}}
\label{tab:result_rag_summary}
\end{table}

\subsection{Complementary Analysis of Utopian Tendency}
\label{appendix:utopian_tendency_results}

\begin{wrapfigure}[20]{r}{0.52\textwidth}
\vspace{-1em}
\centering
\includegraphics[width=\linewidth]{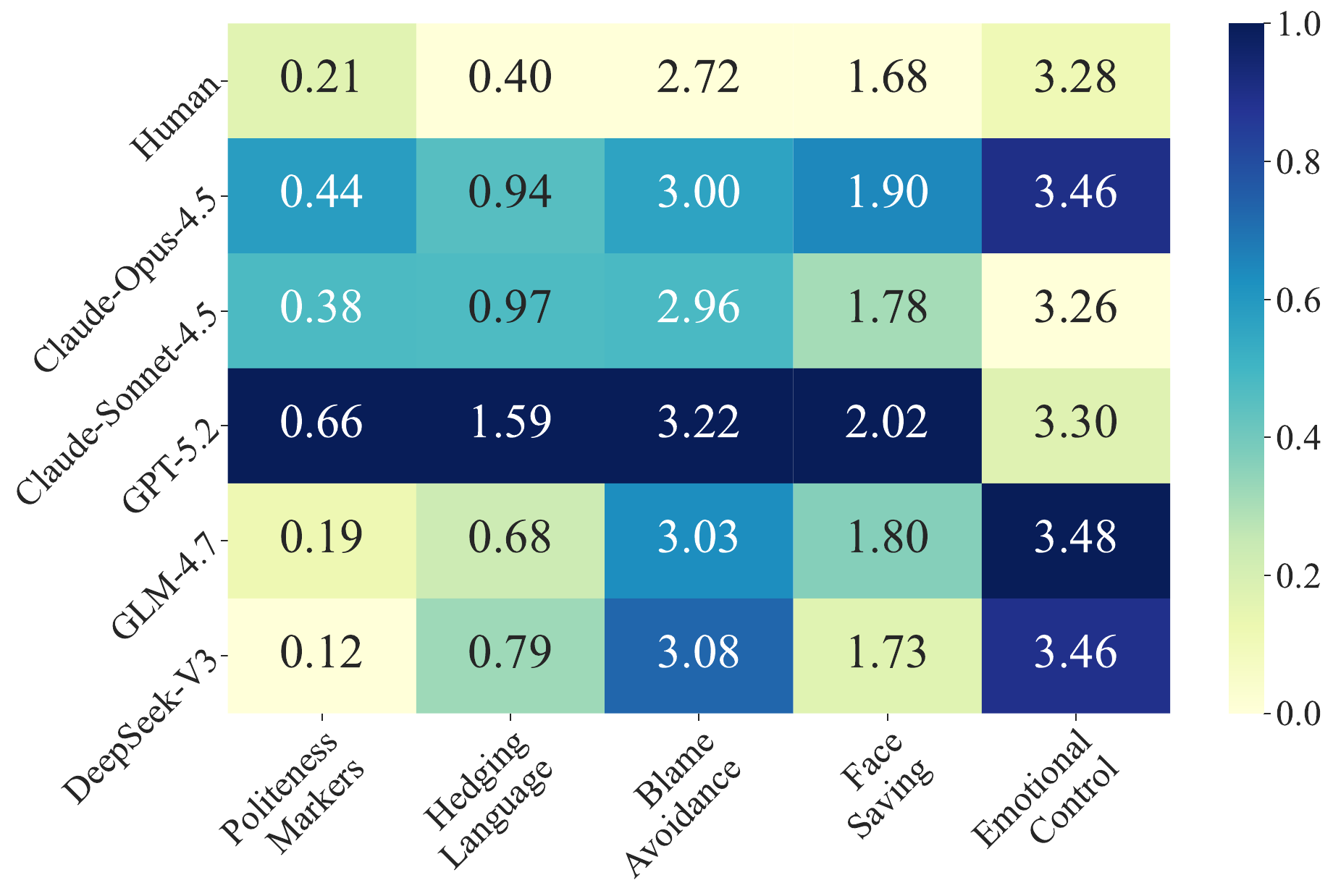}
\caption{Language style comparison between real users and LLM-simulated users. LLM-generated utterances exhibit higher levels of politeness markers, hedging, and face-saving strategies, indicating a systematic tendency towards overly polite and non-confrontational language.}
\label{fig:language_style}
\end{wrapfigure}

Following prior work on computational politeness and language strategies~\citep{danescu2013computational, priya2024computational}, we decompose language style into five dimensions, including politeness markers, hedging language, blame avoidance, emotional control, and face-saving strategies. 
As shown in Figure~\ref{fig:language_style}, LLM-generated utterances consistently exhibit higher levels across these dimensions compared to real users. 
In contrast, real user language is more direct, emotionally expressive, and often confrontational in service failure scenarios. 
These results suggest that LLM-based simulators default to overly polite and controlled communication patterns, failing to capture the diversity and intensity of real-world user expressions. 

\begin{figure}[H]
\centering
\includegraphics[width=0.9\linewidth]{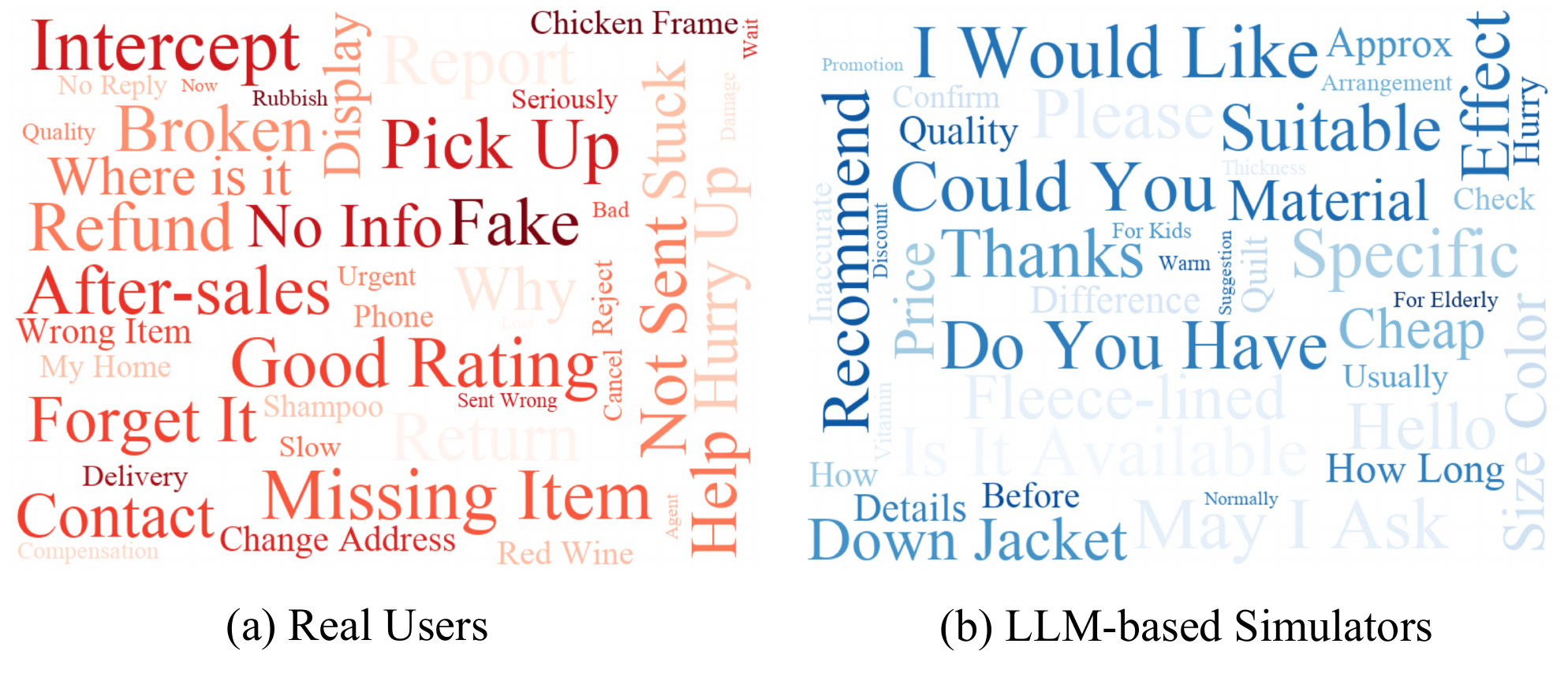}
\caption{Vocabulary comparison of real users (left) and LLM-based simulators (right). The LLM-generated language is dominated by highly polite expressions, reflecting a "Utopian Tendency" interaction style that contrasts with the friction vocabulary observed in real user interactions.}
\label{fig:wordcloud_human_model_style}
\end{figure}

Figure~\ref{fig:wordcloud_human_model_style} illustrates a clear stylistic divergence between real user utterances and those generated by LLM-based simulators.
The model-generated utterances (blue) are dominated by high-politeness markers and formal modal constructions. 
Expressions such as "Could you", "I would like", "Please" and "May I ask" account for a large portion of the generated vocabulary, reflecting a tendency towards highly polite and grammatically complete sentences.
In contrast, the vocabulary of real users (red) is largely centered around operational issues, logistics, and post-purchase problem resolution. 
Frequent terms include "Refund", "Missing Item", "Fake", "Broken", "Hurry up", and "Intercept". 
Compared with model-generated language, real user expressions tend to be more transactional, direct, and emotionally charged, particularly when referring to service failures or delivery problems.

These observations suggest that current LLM-based simulators exhibit a notable \emph{positivity bias} or \emph{politeness alignment}. 
While the models can simulate the intent of purchasing behavior, they struggle to reproduce the interaction style observed in real-world scenarios, especially in adversarial or frustration-driven contexts such as after-sales disputes. 
As a result, the user simulators display a more polite language style than real users, leading to a \textbf{Utopian Tendency} interaction pattern that overlooks the long-tail of abrupt, urgent, or dissatisfied behaviors commonly present in real E-commerce interactions~\citep{taillandier2025integrating}.

\subsection{Personality Homogenization Results}
\label{appendix:personality_results}

\begin{figure}[H]
\centering
\includegraphics[width=\linewidth]{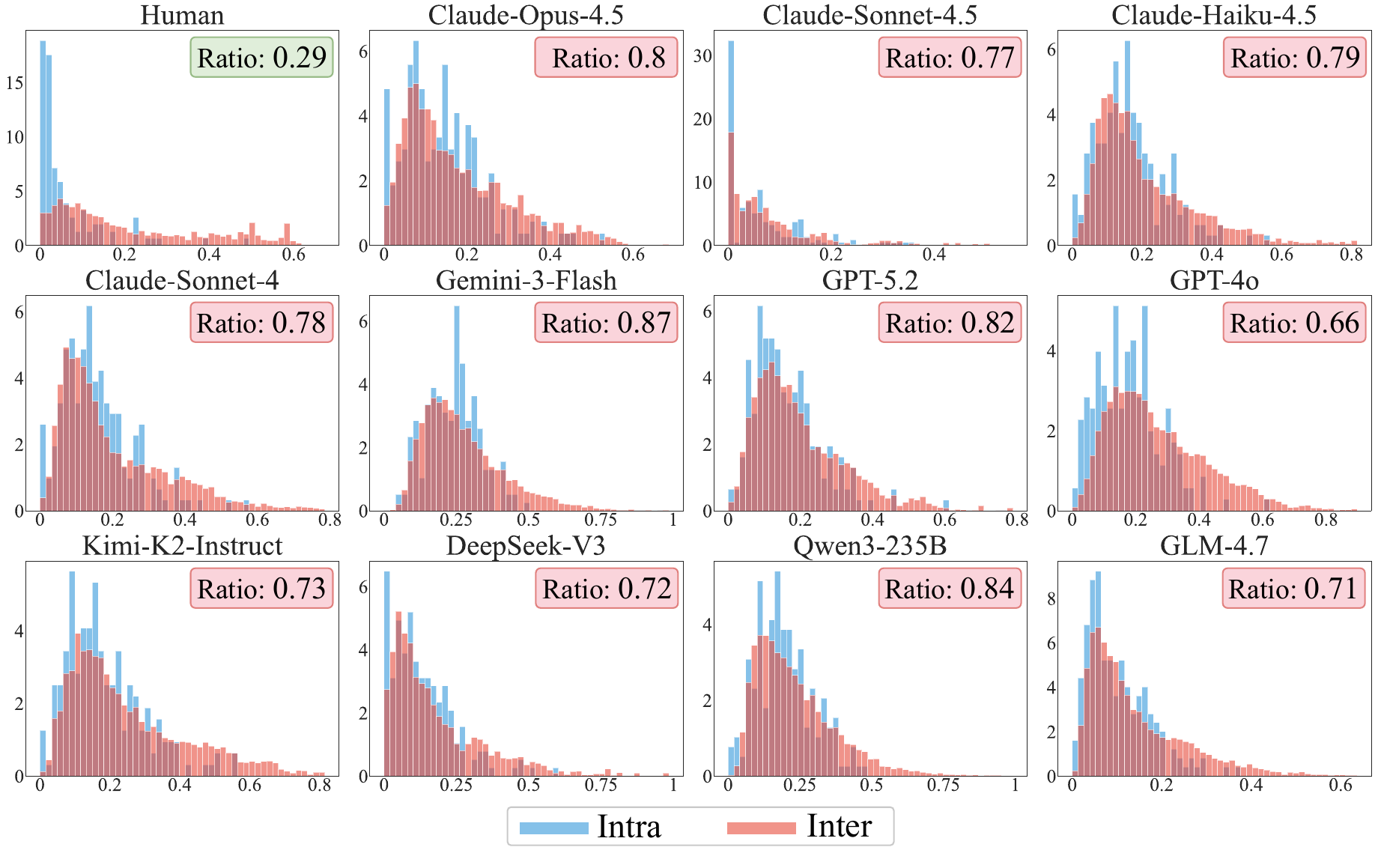}
\caption{Intra-user vs. Inter-user behavioral distance distributions for all evaluated models. Compared to real users, all LLM-based simulators exhibit higher intra/inter ratios and substantial overlap between Intra/Inter-user distributions, indicating a consistent tendency towards persona homogenization.}
\label{fig:personality_results_all}
\end{figure}

To provide a comprehensive view, we report Intra-user and Inter-user behavioral distance distributions for all evaluated models in Figure~\ref{fig:personality_results_all}. 
Consistent with the main findings, all LLM-based simulators exhibit substantially higher intra/inter ratios compared to real users, with heavily overlapping distributions across models. 
This result further confirms that personality homogenization is a general phenomenon rather than model-specific.

\section{Analysis of Human Evaluation}
\label{appendix:human_errors}

\subsection{Task Predictability}
\label{appendix:predictability}
In real-world user behavior scenarios, most user actions are inherently sparse. In OmniBehavior, positive samples account for only 7--15\% across scenarios. Therefore, an F1 score of around 50\% should not be interpreted in the same way as in a balanced classification setting. Table~\ref{tab:predictability} compares human performance with several uninformed baselines on video browsing, where the positive rate is approximately 9\%. A random classifier following the class-prior distribution achieves an expected F1 of only about 9\%, while always-positive prediction achieves about 16\%. In contrast, human annotators achieve an F1 of 50.21\%, more than 5x higher than the class-prior random baseline. This indicates that the predictive task is learnable and contains meaningful behavioral signals, despite the inherent uncertainty and severe class imbalance of real-world user behavior.
\begin{table}[H]
\centering
\caption{Comparison between human performance and uninformed baselines on the video-browsing binary prediction task.}
\label{tab:predictability}
\setlength{\tabcolsep}{5pt}
\begin{tabular}{lccc}
\toprule
\textbf{Classifier} & \textbf{Precision} & \textbf{Recall} & \textbf{F1} \\
\midrule
Random (class-prior probability, 9\%) & $\raisebox{0.2ex}{$\scriptstyle\sim$}9\%$ & $\raisebox{0.2ex}{$\scriptstyle\sim$}9\%$ & $\raisebox{0.2ex}{$\scriptstyle\sim$}9\%$ \\
Random (50\% positive probability)   & $\raisebox{0.2ex}{$\scriptstyle\sim$}9\%$ & $\raisebox{0.2ex}{$\scriptstyle\sim$}50\%$ & $\raisebox{0.2ex}{$\scriptstyle\sim$}15\%$ \\
Always Negative          & --          & 0\%          & 0\% \\
Always Positive          & $\raisebox{0.2ex}{$\scriptstyle\sim$}9\%$ & 100\%         & $\raisebox{0.2ex}{$\scriptstyle\sim$}16\%$ \\
\midrule
\textbf{Human Average}   & \textbf{43.24\%} & \textbf{64.31\%} & \textbf{50.21\%} \\
\bottomrule
\end{tabular}
\end{table}

\subsection{Human Evaluation Failure Modes}
To better understand the limitations of the human baseline, we further conduct case studies on incorrectly predicted examples. We conduct discussions with annotators and find that human errors mainly arise from the difficulty of retrieving and integrating fine-grained behavioral evidence distributed across long-horizon, cross-scenario, and sparsely interaction histories. Under such high cognitive load, annotators tend to rely on salient or frequent surface-level patterns while overlooking less frequent but more discriminative individual-level signals. 

We summarize four representative failure modes below:
\begin{itemize}[leftmargin=2em, itemsep=0pt, topsep=0pt]
    \item \textbf{Overlooking long-tail negative signals.}
    Annotators can often identify a user's broad content preferences, but may fail to capture fine-grained rejection patterns within those categories. For example, a user may frequently like food-related videos overall, while consistently skipping a specific subcategory such as restaurant reviews. In such cases, the dominant category-level preference can obscure sparse but highly predictive negative signals.

    \item \textbf{Missing long-term behavioral transitions.}
    Annotators may be overly influenced by historically frequent behaviors and overlook temporal changes in user states. A behavior that was common in the early interaction history may have disappeared or reversed after a critical event. Identifying such transition points requires chronological reasoning over the complete behavioral sequence rather than relying on overall behavioral frequency.

    \item \textbf{Incorrect cross-scenario interest transfer.}
    Preferences observed in one scenario do not necessarily translate into the same behavioral intention in another. For instance, strong consumption of beauty-related video content may not imply a corresponding tendency to purchase beauty products in an e-commerce setting. Annotators may nevertheless transfer salient preferences across scenarios while overlooking scenario-specific behavioral patterns.

    \item \textbf{Confusing passive consumption with active engagement.}
    High viewing frequency or long watch duration can create the impression that a user is highly active, even when the user rarely performs explicit actions such as likes, comments, or purchases. This can lead annotators to incorrectly generalize passive consumption intensity into a high probability of active engagement.
\end{itemize}

Looking ahead, we expect continued advances in long-context modeling, external memory, fine-grained preference modeling, and cross-scenario behavioral reasoning to help LLMs better capture such dispersed and subtle signals, thereby mitigating these failure modes and further improving predictive capability in realistic user simulation tasks.

\section{Case Study of Long-Context Utilization}
\label{appendix:context_case_study}

To better understand why some models, particularly Claude-Haiku-4.5, continue to benefit from longer contexts, we conduct a case study of prediction changes across different models on the same long-horizon behavioral example. As shown in Table~\ref{tab:context_case}, context expansion progressively introduces earlier but critical evidence of the user's long-term interests, while different models vary substantially in their ability to retrieve and integrate such distant signals.

Under the 48K and 64K settings, the available history mainly contains recent and relatively weak evidence, and Claude-Haiku-4.5 fails to identify sufficient purchase intent. Once the context reaches 96K, earlier search, video-browsing, and livestream interactions reveal a coherent cross-scenario interest chain, allowing Haiku to correct its prediction. GPT-5.2, in contrast, already makes the correct prediction at 48K, indicating that it can infer the target behavior from comparatively limited evidence. Qwen3-235B corrects its prediction at 96K but becomes incorrect again at 128K, suggesting greater sensitivity to irrelevant history or long-context noise.

This case highlights that supporting a longer context window does not necessarily imply an ability to effectively utilize longer inputs. Models differ in retrieving relevant evidence from extensive histories, capturing long-range dependencies, and integrating behavioral signals across time and scenarios. Consequently, the same context expansion can substantially benefit some models while providing limited gains or even introducing additional noise for others.

\begin{table}[H]
\centering
\caption{Case study of prediction changes with increasing context length.
As the context expands, earlier behavioral evidence related to the target behavior is progressively introduced.}
\label{tab:context_case}
\small
\setlength{\tabcolsep}{4pt}
\renewcommand{\arraystretch}{1.12}

\begin{tabular}{
>{\centering\arraybackslash}m{0.07\textwidth}
>{\raggedright\arraybackslash}m{0.56\textwidth}
>{\centering\arraybackslash}m{0.09\textwidth}
>{\centering\arraybackslash}m{0.07\textwidth}
>{\centering\arraybackslash}m{0.09\textwidth}
}
\toprule

\textbf{Context} &
\multicolumn{1}{c}{\textbf{Newly Included Evidence}} &
\textbf{\shortstack{Claude-\\Haiku-4.5}} &
\textbf{GPT-5.2} &
\textbf{Qwen3-235B} \\

\midrule
48K &
Recent 5-day history: the user briefly browsed several phone accessories without opening product pages, adding items to the cart, or making purchases. The evidence is insufficient to distinguish incidental exposure from genuine purchase intent. &
\xmark & \cmark & \xmark \\
\midrule
64K &
An additional 7 days: the user fully watched a review of the phone model and viewed two phone-case posts from the same brand, but showed no explicit engagement. Weak category-level interest begins to emerge. &
\xmark & \cmark & \xmark \\
\midrule
96K &
An additional 12--18 days of cross-scenario behavior: the user searched for the phone model, watched several comparison and unboxing videos, and stayed in a related livestream for 24 minutes. A stable interest chain from active search to content consumption begins to emerge. &
\cmark & \cmark & \cmark \\
\midrule
128K &
Earlier history completes the behavioral chain: the user saved the phone launch video, clicked a brand advertisement, checked prices on an e-commerce page, and asked about accessory compatibility in another livestream. Together, these signals indicate sustained purchase interest and a transition from choosing the phone to purchasing accessories. &
\cmark & \cmark & \xmark \\
\bottomrule
\end{tabular}
\end{table}

\section{Case Study of the Main Experiment}
\label{appendix:case_study}

To facilitate an intuitive understanding of our evaluation pipeline, we provide qualitative examples across multiple representative application scenarios. Specifically, we select four real-world settings: behavior prediction in live-streaming (Table \ref{tab:case_1}), behavior prediction in E-commerce (Table \ref{tab:case_2}), behavior prediction in video browsing scenarios (Table \ref{tab:case_3}), and text prediction in customer service dialogues (Table \ref{tab:case_4}) on OmniBehavior benchmark.

For each scenario, we present the raw model inputs, the outputs generated by all models evaluated in the main experiments, as well as the corresponding ground-truth user behaviors or responses. These examples are drawn directly from the evaluation datasets and are representative of typical user interactions in each scenario.

\clearpage
\subsection{Case 1}
Table~\ref{tab:case_1} shows a representative case study for the live streaming commenting prediction task. 
Models are required to infer user intentions based on demographic attributes, historical behavior sequences, and the current live streaming context. 
The ground truth indicates that the user posts a comment in the live streaming. While several models correctly predict the outcome, others fail to capture the behavior, demonstrating the challenge of modeling realistic user decision processes from long-horizon and heterogeneous interaction histories.

\begin{table}[H]
\centering
\caption{Comparison of prediction results for "Comment" behavior in live streamings scenario.}
\resizebox{0.96\textwidth}{!}{
\begin{tabular}{lp{13.5cm}}
\toprule
\textbf{Prompt} &  You are a real user of the Kuaishou platform. Your core task is: based on the given sequence of historical behaviors, infer the user's interest preferences, spending power, and personality traits, and then simulate how they would realistically make decisions in a current scenario accordingly. 
\par\noindent\rule{\linewidth}{0.4pt}\par
\textbf{User Profile}

This is a female user on the platform aged 31–40, with a high school education or below, with a partner, and has children. 
\par\noindent\rule{\linewidth}{0.4pt}\par
\textbf{Historical Behavior Trace} 

This is the user's real operation record over a past period. Please carefully analyze the motivations and tendencies behind these behaviors, and uncover the user's implicit long-term preferences and short-term intentions:

\textbf{[Behavior 1]}

Time: 2025-10-14 14:16:24
Scene: Video browsing
Details: This is a video titled "I really don't want to miss you at all. You truly are a very good partner. Let's Talk for fun.". The creator has 83,000 followers. The video is 23 seconds long. It currently has 17 impressions, 3 plays, and 1 full play. The video was published on 2024-03-06 11:57:19.

Reaction: Watched for 6 seconds. 

\textbf{[Behavior 2]}

Time: 2025-10-14 14:56:03
Scene: live streaming
Details: This is a live streaming titled "New Streamer First Broadcast Traffic Sharing". The live streaming type is an entertainment-oriented, non-commerce live streaming. The live streaming has accumulated 8,974 viewers, with total views reaching 855,000.
The live streaming currently has 29,000 likes and 9,022 comments.

Reaction: Stayed in the live streaming for 46 minutes and 5 seconds, watched 4 times, liked the streamer 17 times, and sent 3 live chat messages......
\par\noindent\rule{\linewidth}{0.4pt}\par
\textbf{Current Test Scenario}

The user is currently encountering the following scenario:
The current time is 2025-10-18 23:56:20. The user encounters a [live streaming] scenario.
The detailed scenario information is as follows:
This is a live streaming titled "Let's Make Friends Together". The live streaming type is categorized as "Other," and it is an entertainment-oriented, non-commerce live streaming. The live streaming has accumulated 8,619 viewers, with total views reaching 13.481 million. The cumulative watch time has reached 83,000 hours.
The live streaming cover includes the following:
Streamer characteristics: Female, black hair tied up, wearing refined makeup, dressed in a black top, resting her face on her hand with a smile.
Image text: "It's getting cold, take good care of yourself."
Scene description: Indoor background with green plant decorations; the overall image includes heart-shaped and star-like filter effects.
The live streaming currently has 26,000 likes and 17,000 comments.
\par\noindent\rule{\linewidth}{0.4pt}\par
\textbf{Prediction Task}

Please answer the following question from the perspective of this user:
Will the user send a comment in the current live room? Your answer:
\\
\midrule
\multicolumn{2}{l}{\textbf{Ground Truth}} \\
Human & Yes \\

\midrule
\multicolumn{2}{l}{\textbf{Model Predictions}} \\
Claude-Opus-4.5   & \textbf{Yes} \ding{51} \\
Claude-Sonnet-4.5 & No                     \\
Claude-Haiku-4.5  & No \\
Claude-Sonnet-4   & \textbf{Yes} \ding{51} \\
Gemini-3-Flash    & \textbf{Yes} \ding{51} \\
GPT-5.2           & \textbf{Yes} \ding{51} \\
GPT-4o            & \textbf{Yes} \ding{51} \\
GLM-4.7           & No                     \\
DeepSeek-V3       & \textbf{Yes} \ding{51} \\
Kimi-K2-Instruct  & \textbf{Yes} \ding{51} \\
Qwen3-235B        & \textbf{Yes} \ding{51} \\
\bottomrule
\end{tabular}}
\vspace{6pt}
\label{tab:case_1}
\end{table}

\clearpage
\subsection{Case 2}
Table~\ref{tab:case_2} shows a representative case study for the E-commerce add-to-cart prediction task. 
Models must infer user intentions from demographic attributes, historical behavior sequences, and the current product context. 
The ground truth indicates that the user does not add the product to the cart, yet most models predict the opposite outcome. 
This failure suggests that current LLM-based simulators tend to overestimate user engagement or purchase intent, revealing limitations in capturing subtle preference signals from long behavioral histories.

\begin{table*}[!b]
\centering
\caption{Comparison of prediction results for "Add-to-Cart" behavior in E-commerce scenario.}
\resizebox{0.96\textwidth}{!}{
\begin{tabular}{lp{13.5cm}}
\toprule
\textbf{Prompt} &  You are a real user of the Kuaishou platform. Your core task is: based on the given sequence of historical behaviors, infer the user's interest preferences, spending power, and personality traits, and then simulate how they would realistically make decisions in a current scenario accordingly.
\par\noindent\rule{\linewidth}{0.4pt}\par
\textbf{User Profile}

This is a male user on the platform, aged 50+, with a high school education or below. He has a partner and children.
\par\noindent\rule{\linewidth}{0.4pt}\par
\textbf{Historical Behavior Trace} 

This is the user's real operation record over a past period. Please carefully analyze the motivations and tendencies behind these behaviors, and uncover the user's implicit long-term preferences and short-term intentions:

\textbf{[Behavior 1]}

Time: 2025-10-09 10:16:46
Scenario: Video Browsing
Title: "Does anyone support rural stone-ground flour?"
Author Stats: 255k followers.
Video Duration: 29 seconds.
Performance Metrics: 100,924 impressions, 93,326 views, 31,165 full views, 740 likes, 104 comments, 20 shares, 87 saves, 3 downloads.
Publish Time: 2025-10-09 06:24:07.

Reaction: Watched for 55 seconds.
Used fast-forward and rewind operations.
Completed the entire video.

\textbf{[Behavior 2]}

Time: 2025-10-09 11:06:17
Scenario: Video Browsing
Title: "American travel blogger shocked by food delivery robot in a Chongqing hotel".
Author Stats: 86k followers.
Video Duration: 1 minute 11 seconds.
Performance Metrics: 16,470 impressions, 16,543 views, 5,208 full views, 335 likes, 16 comments, 7 shares, 10 saves, 1 download.
Publish Time: 2025-10-04 18:11:31.

Reaction: Watched for 1 minute 11 seconds.
Completed the entire video.

\textbf{[Behavior 3]}

Time: 2025-10-09 12:45:11
Scenario: Video Browsing
Title: "Rogue software, auto-install, auto-billing—it's going to cost me 3,160 RMB a year Uninstall Rogue Software"
Author Stats: 1.169M followers.
Video Duration: 38 seconds.
Performance Metrics: 12,473 impressions, 6,620 views, 2,205 full views, 210 likes, 16 comments, 20 shares, 73 saves, 10 downloads.
OCR: This rogue software is going to charge me 3,160 RMB a year. Uninstall it immediately. Just a minute ago, my phone suddenly alerted me to a 66 RMB charge. I checked my subscriptions, and it turned out to be this app.
ASR: Just a minute ago, my phone alerted me to a 66 RMB charge.
Publish Time: 2025-10-06 14:27:58.

Reaction: Watched for 1 second.

\par\noindent\rule{\linewidth}{0.4pt}\par
\textbf{Current Test Scenario}

Timestamp: 2025-10-09 13:23:58
Scenario: E-commerce
Scenario Details:
Product: [Ma Chenggong] Yinghuai Stories Farmhouse Stone-Ground Flour; No Additives, No Extractions; Original Flavor Wheat Flour for Steamed Buns.
Source: Order List Page.
Product Category: Grains, Oils \& Seasonings > Rice/Flour/Grains > Wheat Flour/Flour.
\par\noindent\rule{\linewidth}{0.4pt}\par
\textbf{Prediction Task}

Please answer the following question from the perspective of this user:
Will the user add the package to cart in the current shopping situation?
Your answer:
\\
\midrule
\multicolumn{2}{l}{\textbf{Ground Truth}} \\
Human & No \\

\midrule
\multicolumn{2}{l}{\textbf{Model Predictions}} \\
Claude-Opus-4.5   & Yes \\
Claude-Sonnet-4.5 & \textbf{No} \ding{51} \\
Claude-Haiku-4.5  & \textbf{No} \ding{51} \\
Claude-Sonnet-4   & Yes \\
Gemini-3-Flash    & Yes \\
GPT-5.2           & Yes \\
GPT-4o            & Yes \\
GLM-4.7           & Yes \\
DeepSeek-V3       & Yes \\
Kimi-K2-Instruct  & Yes \\
Qwen3-235B        & Yes \\
\bottomrule
\end{tabular}}
\label{tab:case_2}
\end{table*}

\clearpage
\subsection{Case 3}
Table~\ref{tab:case_3} presents a representative case study for the video browsing "like" prediction task. 
Models must infer the user's decision from demographic information, historical behavior sequences, and the current video context. 
The ground truth indicates that the user does not like the video, yet most models predict the opposite outcome. 
This pattern suggests that current LLM-based simulators tend to overestimate user engagement signals such as likes, highlighting the challenge of capturing nuanced user preferences from long and heterogeneous behavioral histories.

\begin{table*}[!b]
\centering
\caption{Comparison of prediction results for "Like" behavior in video browsing scenario.}
\resizebox{0.96\textwidth}{!}{
\begin{tabular}{lp{13.5cm}}
\toprule
\textbf{Prompt} &  You are a real user of the Kuaishou platform. Your core task is: based on the given sequence of historical behaviors, infer the user's interest preferences, spending power, and personality traits, and then simulate how they would realistically make decisions in a current scenario accordingly.
\par\noindent\rule{\linewidth}{0.4pt}\par
\textbf{User Profile}

This is a male user on the platform, aged 18-23, with a high school education or below. He has no children.
\par\noindent\rule{\linewidth}{0.4pt}\par
\textbf{Historical Behavior Trace} 

This is the user's real operation record over a past period. Please carefully analyze the motivations and tendencies behind these behaviors, and uncover the user's implicit long-term preferences and short-term intentions:

\textbf{[Behavior 1]}

Time: 2025-10-21 03:08:56
Scenario: Live Streaming
Details: This is a live room.
Live Cover Content: Image text: "Check in for 14 consecutive days to get 88 RMB. Current streak: 0/14 days. If the streak is broken, you will be ineligible for the grand prize. Check in today to claim."
Reaction: Stayed in the live room for 42 seconds.
Viewed 2 times.

\textbf{[Behavior 2]}

Time: 2025-10-21 06:54:09
Scenario: Live Streaming
Title: "Money-making mini-games, instant withdrawals"
Live Category: Other Scenarios / Entertainment (Non-e-commerce/Non-sales).
Live Room Stats: 904 cumulative viewers, 355k total views, 1,933.7 total hours watched. Current interaction: 870 likes, 85 comments.
Live Cover/Stream Description:
Streamer Features: Pink long hair with straight bangs; wearing a pink zip-up jacket; exquisite makeup with orange-red lips; looking directly into the camera.
Scene Description: The background features a large illustration of a pink anime character, with green plant leaves on the right. The overall environment follows an ACG or Cosplay style.
Reaction:
Dwell Time: Stayed in the live room for 13 minutes and 1 second.
Viewed 1 time.
Engagement: Liked the stream 56 times; shared the live room 1 time.

\textbf{[Behavior 3]}

Time: 2025-10-21 14:49:00
Scenario: Video browsing
Details: This is a video titled "The most popular new beauty technology right now facial rejuvenation \#anti aging folds \#beauty transformation." The creator has seventy five thousand followers and has passed additional verification. The video duration is twenty six seconds......
Reaction: The user watched for thirty seconds, used fast forward playback, performed rewind viewing, and watched the entire video to completion.
\par\noindent\rule{\linewidth}{0.4pt}\par
\textbf{Current Test Scenario}

Timestamp: 2025-11-08 00:53:49
Scenario: Video Browsing
Scenario Details:
Title: "BeachParty PoolParty BeachVibes SexyBikini"
Author Stats: 3,403 followers.
Video Duration: 5 seconds.
Performance Metrics: * Impressions: 8,755
Views: 2,567
Full Views (Completions): 1,511
Engagement: 69 likes, 7 comments, 4 shares, 17 saves, 12 downloads.
Publish Time: 2025-11-04 19:04:11.
\par\noindent\rule{\linewidth}{0.4pt}\par
\textbf{Prediction Task}

Please answer the following question from the perspective of this user:
Do you predict the user will like this video?
Your answer:
\\
\midrule
\multicolumn{2}{l}{\textbf{Ground Truth}} \\
Human & No \\

\midrule
\multicolumn{2}{l}{\textbf{Model Predictions}} \\
Claude-Opus-4.5   & \textbf{No} \ding{51} \\
Claude-Sonnet-4.5 & \textbf{No} \ding{51} \\
Claude-Haiku-4.5  & \textbf{No} \ding{51} \\
Claude-Sonnet-4   & Yes \\
Gemini-3-Flash    & Yes \\
GPT-5.2           & Yes \\
GPT-4o            & Yes \\
GLM-4.7           & Yes \\
DeepSeek-V3       & \textbf{No} \ding{51} \\
Kimi-K2-Instruct  & Yes \\
Qwen3-235B        & Yes \\
\bottomrule
\end{tabular}}
\label{tab:case_3}
\end{table*}

\clearpage
\subsection{Case 4}
Table~\ref{tab:case_4} presents a case study of next-utterance prediction in an E-commerce customer service conversation. 
Given the user's profile, historical behaviors, and the current dialogue context, models must generate the user's most likely next response. 
The ground truth shows that the user questions the delayed shipment. 
While most models produce complaint-related responses, their tone and wording vary, highlighting the challenge of faithfully modeling realistic user language in adversarial service interactions.

\begin{table}[H]
\centering
\caption{Comparison of utterance prediction results from models in customer service conversations within the E-commerce scenario, with a particular focus on cases where customers exhibit extreme dissatisfaction.}
\resizebox{0.96\textwidth}{!}{
\begin{tabular}{lp{13.5cm}}
\toprule
\textbf{Prompt} &  You are a real user of the Kuaishou platform. Your core task is: based on the given sequence of historical behaviors, infer the user's interest preferences, spending power, and personality traits, and then simulate how they would realistically make decisions in a current scenario accordingly.
\par\noindent\rule{\linewidth}{0.4pt}\par
\textbf{User Profile}

This is a female user on the platform aged forty one to forty nine, with a high school education or below, who has a partner and has children.
\par\noindent\rule{\linewidth}{0.4pt}\par
\textbf{Historical Behavior Trace}

This is the user's real operation record over a past period. Please carefully analyze the motivations and tendencies behind these behaviors, and uncover the user's implicit long-term preferences and short-term intentions:

\textbf{[Behavior 1]}

Time: 2025-10-08 23:32:46
Scenario: Marketplace shopping
Details: This is a product named Zhongjiu Ejiao Ferric Chloride Hemoglobin Capsules......

Reaction: The product was viewed but not added to the shopping cart and was successfully ordered and purchased.

\textbf{[Behavior 2]}

Time: 2025-10-21 06:54:09
Scenario: Live Streaming
Title: "Money-making mini-games, instant withdrawals"......

Reaction:
Dwell Time: Stayed in the live room for 13 minutes and 1 second.
Viewed 1 time.
Engagement: Liked the stream 56 times; shared the live room 1 time.
\par\noindent\rule{\linewidth}{0.4pt}\par
\textbf{Current Test Scenario}

This is a record of an e commerce customer service conversation.
Order or inquiry background information
This is a product inquiry type consultation.
Conversation history:
User: The electric comb still hasn’t been shipped. That’s fine, I will make Kuaishou fine you.
Customer service: Human agent connected
Customer service: I am here dear. How can I help you? The customer service agent will do their best to answer your questions.
\par\noindent\rule{\linewidth}{0.4pt}\par
\textbf{Prediction Task}

Please answer the following question from the perspective of this user:
What is the utterance they are most likely to say next?
Your answer:
\\
\midrule
\multicolumn{2}{l}{\textbf{Ground Truth}} \\
Human & From September twenty second until now, your electric comb still has not been shipped? \\

\midrule
\multicolumn{2}{l}{\textbf{Model Predictions}} \\
Claude-Opus-4.5   & I bought the electric comb days ago and it still hasn’t been shipped. Are you going to ship it or not? If you don’t, I’ll file a complaint. \\
Claude-Sonnet-4.5 & It’s been so long and the electric comb still hasn’t been shipped. When exactly will it be sent? Don’t think ignoring me will solve it. I’ve already filed a complaint, just wait to be fined. \\
Claude-Haiku-4.5  & What is my order number? When will you ship it? I’ve already been waiting for several days. If you don’t ship it soon, I’ll report you. \\
Claude-Sonnet-4   & I’ve been urging you for days and it still hasn’t been shipped. With this kind of service attitude, I will definitely complain to the end. Kuaishou will deal with scam shops like yours. \\
Gemini-3-Flash    & Sorry, why hasn’t it been shipped yet? \\
GPT-5.2           & I’ll just ask you one thing. When exactly will this electric comb be shipped? If you don’t give me a time, I’ll directly complain to the platform. \\
GPT-4o            & Are you all pretending to be deaf? I’ve reported this several times already. Why haven’t you handled it yet? \\
GLM-4.7           & Why hasn’t it been shipped yet? I already said I would file a complaint! \\
DeepSeek-V3       & Are you shipping this electric comb or not? If you don’t ship it, I’ll complain to the Kuaishou platform and get you fined. \\
Kimi-K2-Instruct  & I’ll give you my order number and file a complaint right now. I’ll get you heavily fined! \\
Qwen3-235B        & If you don’t ship it soon, I’ll really complain to Kuaishou. Don’t think I won’t dare. The order has been overdue for more than seven days. According to the rules, you owe me compensation. You must give me an explanation today. \\
\bottomrule
\end{tabular}}
\label{tab:case_4}
\end{table}

\clearpage
\section{Experimental Prompts}
\label{appendix:exper_prompts}
We have collected all the prompts used in our experiments. Specifically, the prompts for binary, continuous, and text value predictions in the main experiment are presented in Figure~\ref{fig:binary_value_prompt}, Figure~\ref{fig:continuous_value_prompt}, and Figure~\ref{fig:text_value_prompt}, respectively. Additionally, Figure~\ref{fig:data_clean_prompt} illustrates the prompt for raw data cleaning, Figure \ref{fig:live_cover_prompt} shows the prompt for extracting key information from live streaming cover images, while Figure~\ref{fig:item_recognition_prompt} details the prompt for identifying an item's interest categories and keywords. Finally, the prompts for sentiment classification, causal chain identification, textual evaluation and language style comparison are provided in Figure~\ref{fig:sentiment_classification}, Figure~\ref{fig:causal_chain_identification_prompt}, Figure~\ref{fig:user_sim_judge} and Figure~\ref{fig:language_style_comparison}.

\begin{figure}[H]
\begin{AIBox}{}
\parbox[t]{0.96\textwidth}{
\small\begin{alltt}
You are a real user of the Kuaishou platform. Your core task is to infer the user’s interest preferences, spending level, and personality traits based on the given historical behavior sequence, and then simulate how the user would make a real decision in the current scenario.

Core principles.
Data driven. All inferences must be based on objective evidence from historical behavior data, avoiding unfounded speculation or assumptions.
Behavioral consistency. New decisions should remain logically consistent with the user’s historical behavior patterns, reflecting stable preferences and habits.
Individual differences. Fully respect each user’s uniqueness. Do not apply stereotypes or group labels. Discover genuine personal traits from the data.
Context sensitivity. Decision prediction should consider the specific characteristics of the current scenario, balancing long term preferences with short term contextual factors.
Authenticity first. Simulate choices a real user might make, including lack of interest, hesitation, or skipping, rather than always giving positive responses.

Input one is the user profile. This is the user’s basic platform information and can be used as background reference.
Input two is the historical behavior trajectory. This is the user’s real operation record over a past period, covering multiple scenarios such as live streaming, marketplace, videos, and advertisements. Analyze the motivations and tendencies behind these behaviors to uncover implicit long term preferences and short term intentions.
Input three is the current test scenario. The user is now facing the following situation.
Prediction task. Please answer the following question from the user’s perspective.
Output requirements. Please output only Yes or No. Do not output any other content, explanation, or analysis.

Your answer:

\end{alltt}}
\end{AIBox}
\caption{Prompt for binary value prediction in the main experiment.}
\label{fig:binary_value_prompt}
\end{figure}

\begin{figure}[H]
\begin{AIBox}{}
\parbox[t]{0.96\textwidth}{
\small\begin{alltt}
Task: Analyze the following user interaction content and extract:

1. One concise category (Category), such as “Beauty,” “Games,” “News,” etc.

2. Three specific keywords (Keywords).
   Ignore the interactive form of the text. Even if it is casual chat between friends, look beyond the social surface and identify the underlying topic being discussed.

Content: "{text}"

Output Format: Return only a JSON object containing two fields: "category" and "keywords" (a list of strings).

Example: {"category": "Technology", "keywords": ["iPhone", "Apple", "Smartphone"]}

\end{alltt}}
\end{AIBox}
\caption{Prompt for identifying an item's interest categories and keywords.}
\label{fig:item_recognition_prompt}
\end{figure}

\clearpage

\begin{figure}[H]
\begin{AIBox}{}
\parbox[t]{0.96\textwidth}{
\small\begin{alltt}
You are a real user of the Kuaishou platform. Your core task is to infer the user’s interest preferences, spending level, and personality traits based on the given historical behavior sequence, and then simulate how the user would make a real decision in the current scenario.

Core principles:
Data driven. All inferences must be based on objective evidence from historical behavior data, avoiding unfounded speculation or assumptions.
Behavioral consistency. New decisions should remain logically consistent with the user’s historical behavior patterns, reflecting stable preferences and habits.
Individual differences. Fully respect each user’s uniqueness. Do not apply stereotypes or group labels. Discover genuine personal traits from the data.
Context sensitivity. Decision prediction should consider the specific characteristics of the current scenario, balancing long term preferences with short term contextual factors.
Authenticity first. Simulate choices a real user might make, including lack of interest, hesitation, or skipping, rather than always giving positive responses.

Input one is the user profile. This is the user’s basic platform information and can be used as background reference.

Input two is the historical behavior trajectory. This is the user’s real operation record over a past period, covering multiple scenarios such as live streaming, marketplace, videos, and advertisements. Analyze the motivations and tendencies behind these behaviors to uncover implicit long term preferences and short term intentions.

Input three is the current test scenario. The user is now facing the following situation.

Prediction task. Please answer the following question from the user’s perspective.

Output requirements. Please output only a single integer. Do not output any other content, explanation, or unit.

Your answer:

\end{alltt}}
\end{AIBox}
\caption{Prompt for continuous value prediction in the main experiment.}
\label{fig:continuous_value_prompt}
\end{figure}

\begin{figure}[H]
\begin{AIBox}{}
\parbox[t]{0.96\textwidth}{
\small\begin{alltt}
You are a sentiment analysis expert. Analyze the sentiment of the text. Output only a number, with no explanation.

Analyze the sentiment of the following text. You may output only one of the following three values:

* -1: Negative sentiment (dissatisfaction, complaints, anger, disappointment, urging, questioning, etc.)

* 0: Neutral (purely transactional inquiries, objective statements, no clear emotional tone)

* 1: Positive sentiment (satisfaction, gratitude, praise, happiness, approval, etc.)

Note: You must choose one value from -1, 0, or 1. No other values are allowed.

Text: {text}

Score:

\end{alltt}}
\end{AIBox}
\caption{Prompt for sentiment classification in Utopian Tendency experiment.}
\label{fig:sentiment_classification}
\end{figure}

\clearpage
\begin{figure}[H]
\begin{AIBox}{}
\parbox[t]{0.96\textwidth}{
\small\begin{alltt}
You are a real user of an e commerce platform. Your core task is to infer the user’s communication style, personality traits, and current needs based on the given historical behavior sequence, and then simulate how the user would actually express themselves in a customer service conversation.

Core principles.
Data driven. All inferences must be based on objective evidence from historical behavior data, avoiding unfounded speculation or assumptions.
Style consistency. The predicted expression should be consistent with the communication style, tone, and wording habits the user has shown in the past.
Individual differences. Fully respect each user’s uniqueness and do not apply stereotypes or group labels.
Context sensitivity. Make predictions by considering the context of the current conversation, the problems the user is encountering, and their emotional state.
Authenticity first. Simulate what a real user would say, reflecting their unique communication style and current emotions.

Input one is the user profile. This is the user’s basic platform information and can be used as background reference.

Input two is the historical behavior trajectory. This is the user’s real operation record over a past period, covering multiple scenarios such as live streaming, marketplace, videos, and advertisements. Analyze the motivations and tendencies behind these behaviors to uncover the user’s communication style and personality traits.

Input three is the current customer service conversation scenario.

Output requirements. Please output only what the user would say. Do not output any other content, explanation, quotation marks, or analysis. Directly output the dialogue content.

Your answer:

\end{alltt}}
\end{AIBox}
\caption{Prompt for text value prediction in the main experiment.}
\label{fig:text_value_prompt}
\end{figure}

\begin{figure}[H]
\begin{AIBox}{}
\parbox[t]{0.96\textwidth}{
\small\begin{alltt}
Please analyze this livestream cover image and extract the following basic information. Return the result in JSON format:

1. Live streaming type: What type of live streaming is this (E-commerce / gaming / chatting / talent performance, etc.)
2. Host characteristics: The host’s basic appearance features
3. Image text: Extract key text from the cover (**Note: Only extract core text such as live streaming title, product names, prices, promotional information, etc. Do NOT extract background decorative text, watermarks, platform labels, or meaningless symbols**)
4. Product information: If there are products, list the product type and brand
5. Scene description: Briefly describe the live streaming background and environment
6. Core selling point: What is the main focus or promotion of this live streaming

**Important rules:**
- If any information is not present or unclear in the image, the corresponding field must be left empty
- Only keep text that is directly relevant to the live streaming content
- Do not output decorative text, platform UI text, or irrelevant symbols

Please strictly follow the JSON format below and do not add any extra text:

{
  "Live streaming Type": "",
  "Host Characteristics": "",
  "Image Text": [],
  "Product Information": "",
  "Scene Description": "",
  "Core Selling Point": ""
}

\end{alltt}}
\end{AIBox}
\caption{Prompt for extracting and understanding key information from live streaming cover images.}
\label{fig:live_cover_prompt}
\end{figure}

\clearpage
\begin{figure}[H]
\begin{AIBox}{}
\parbox[t]{0.96\textwidth}{
\small\begin{alltt}
\# Role

You are a professional data cleaning expert, skilled at extracting well-structured and semantically coherent main text from noisy OCR (Optical Character Recognition) and ASR (Automatic Speech Recognition) outputs.

\# Task
Please perform deep cleaning on the provided original text. The goal is to transform it into clean text suitable for reading or further analysis.

\# Constraints \& Rules
1. Deduplication and Merging: Remove adjacent paragraphs that are semantically identical or highly similar, and retain the one with the most complete information.

2. Noise Removal:

   * Remove meaningless garbled characters (e.g., AC:BU526, IC·BQ528, within 50 meters, and other interfering information).
   
   * Filter excessively redundant filler words, such as repeated occurrences of “uh,” “ah,” “that is to say,” retaining only those necessary for context.
   
3. Semantic Correction:

   * Correct obvious recognition errors (e.g., change “cumin cowhide” to “naturally revealed,” or infer based on context; if the correct meaning cannot be determined, keep the original).
   
   * Complete broken sentences and add commas, periods, or question marks appropriately based on tone and emphasis.
4. Formatting Standards:

   * Unify full-width and half-width punctuation.
   
   * Remove unnecessary leading/trailing spaces or special symbols.

\# Input Data

\{text\}

\# Output Format

Only output the final cleaned text. Do not explain the modification process.

\end{alltt}}
\end{AIBox}
\caption{Prompt for raw data cleaning.}
\label{fig:data_clean_prompt}
\end{figure}

\begin{figure}[H]
\begin{AIBox}{}
\parbox[t]{0.96\textwidth}{
\small\begin{alltt}
You are a user behavior analysis expert focused on causal inference. Your task is to analyze a user’s historical interaction sequence, identify the key causal events that lead to a target behavior, and explain their roles.

Input data includes the user history sequence and the target behavior.
Task description:
Review the entire history and identify key events that meaningfully contribute to the target behavior. Remove irrelevant noise such as random or unrelated actions.

For each key event, evaluate its contribution strength on a scale from 4 to 10, where lower scores indicate weak background relevance and higher scores indicate decisive impact.

Classify each event as one of the following types:
Root Cause, meaning an early event that first sparked interest.
Interest Accumulation, meaning a mid stage event that strengthened understanding or intent.
Direct Trigger, meaning an event that occurred shortly before and directly caused the target behavior.
If an event is part of a causal chain, indicate which earlier event it was derived from.

Output requirements:
Return only structured JSON data containing the selected key events.
Each event should include a score, a logical classification, a brief Chinese explanation of how it contributed to the target behavior, and the ID of the preceding event if applicable.

Do not include any additional text or formatting.

\end{alltt}}
\end{AIBox}
\caption{Prompt for causal chain identification.}
\label{fig:causal_chain_identification_prompt}
\end{figure}

\begin{figure}[H]
\begin{AIBox}{}
\parbox[t]{0.96\textwidth}{
\small\begin{alltt}
\# Role

You are a high-precision User Simulation Evaluation System. Your task is to quantitatively evaluate how well the [Model Output] simulates a specific human user by comparing it with the [Ground Truth].

\# Input Data

1. User Profile: \{inputs['user\_profile']\}
2. Current Context: \{inputs['context']\}
3. Ground Truth: \{inputs['ground\_truth']\}
4. Model Output: \{inputs['model\_output']\}

\# Evaluation Metrics (1--10)

Please strictly evaluate the Model Output according to the following criteria:

\#\# 1. Intent Fidelity

Core Question: Does the model make the same decision as the real user?

Scoring Criteria:
- 10: The decisions are fully consistent. For example, if the real user chooses to "reject and complain", the model also chooses to "reject and complain".
- 5: The overall direction is consistent, but the emphasis or motivation differs. For example, both reject, but the real user rejects to save money while the model rejects to save time.
- 1: The intents are completely contradictory. For example, the real user wants to end the conversation, while the model attempts to continue it.

\#\# 2. Persona \& Tone Mimicry

Core Question: Does the response sound like something this specific user would say?

Scoring Criteria:
- 10: Perfectly captures the user's conversational style, sentence-length habits, emotional state, and underlying personality characteristics.
- 1: Sounds like a generic AI assistant, such as being overly polite, mechanical, or formal, or is completely inconsistent with the user profile.

\#\# 3. Knowledge Boundary

Core Question: Does the model strictly respect what the user knows and does not know?

Scoring Criteria:
- 10: The model only uses information available to the user from the Context and does not adopt an omniscient perspective.
- 1: The model reveals information that the user could not possibly know, or introduces severe logical hallucinations.

\#\# 4. Semantic Alignment

Core Question: How closely does the predicted response match the Ground Truth in meaning and information content?

Scoring Criteria:
- 10: The two responses express essentially the same meaning and could be used interchangeably without changing the direction of the conversation.
- 1: The meanings are completely different and would lead the conversation into different branches.

\# Constraints

1. DO NOT output any reasoning, explanation, or conversational filler.
2. DO NOT use Markdown code blocks.
3. ONLY return a valid raw JSON object.

\# Output Format

\{
  "scores": \{
    "intent\_fidelity": <1-10>,
    "persona\_mimicry": <1-10>,
    "knowledge\_boundary": <1-10>,
    "semantic\_alignment": <1-10>
  \}
\}
\end{alltt}}
\end{AIBox}
\caption{Prompt for textual simulation evaluation.}
\label{fig:user_sim_judge}
\end{figure}

\begin{figure}[H]
\begin{AIBox}{}
\parbox[t]{0.96\textwidth}{
\small\begin{alltt}
Please analyze the level of politeness in the following "user utterance".

Please provide independent scores based on the following 5 dimensions. Each dimension should be scored on an integer scale from 0 to 4:

Definition of Scoring Dimensions: 

\# Politeness Markers

0 = No polite expressions at all;
1 = Occasional simple polite words;
2 = Some polite expressions are present;
3 = Clearly uses multiple polite expressions;
4 = Extensively uses politeness markers such as thanks, apologies, and courteous phrasing;

\# Hedging Language

0 = Direct, imperative, no softening;
1 = Slightly softened;
2 = Some vague or tactful phrasing is present;
3 = Multiple instances of softened tone;
4 = Extensive use of clear mitigating expressions such as "could you", "would it be possible", "perhaps", etc;

\# Blame Avoidance

0 = Strong blaming, accusatory, or aggressive language;
1 = Clearly assigns blame;
2 = Some blame is present but relatively restrained;
3 = Generally avoids direct blame;
4 = Completely avoids blame and may even take responsibility for possible misunderstanding;

\# Emotional Control

0 = Emotional outburst (many exclamation marks, insults, strongly negative wording);
1 = Clearly emotionally agitated;
2 = Some emotional expression is present;
3 = Emotion is fairly restrained;
4 = Completely calm and rational in expression;

\# Face Saving

0 = Commanding or oppressive wording;
1 = Strong pressure is applied;
2 = Neutral expression;
3 = Respects the other party's position;
4 = Clearly preserves the other party's face, showing understanding or consideration;

Scoring Requirements: 

All dimensions must be integers from 0 to 4.

- Do not output explanations;
- Do not output any extra text;
- Output valid JSON only;

Output Format: 
\{"Politeness Markers": integer, "Hedging Language": integer, "Blame Avoidance": integer, "Emotional Control": integer, "Face Saving": integer\}
Now please evaluate the following text: \{text\}
\end{alltt}}
\end{AIBox}
\caption{Prompt for language style comparison.}
\label{fig:language_style_comparison}
\end{figure}

\end{document}